\pdfoutput=1
\documentclass[11pt]{article}

\usepackage[preprint]{acl}

\usepackage{times}
\usepackage{latexsym}
\usepackage[T1]{fontenc}
\usepackage[utf8]{inputenc}
\usepackage{microtype}
\usepackage{inconsolata}
\usepackage{graphicx}
\usepackage{booktabs}
\usepackage{multirow}
\usepackage{makecell}
\usepackage{amssymb}
\usepackage{array}
\usepackage{algorithm}
\usepackage{algpseudocode}

\usepackage{enumitem}
\usepackage{xcolor}
\usepackage{colortbl}
\usepackage{tikz}
\usepackage{float}
\usetikzlibrary{patterns}
\usepackage{arydshln}
\usepackage{soul}
\usepackage{hyperref}
\usepackage{subcaption}
\usepackage[most]{tcolorbox}
\usepackage{microtype}

\tcbuselibrary{skins,listings,breakable}
\definecolor{promptaccent}{HTML}{34506B}
\definecolor{promptback}{HTML}{FAFBFD}
\colorlet{solveraccent}{promptaccent}
\colorlet{solverback}{promptback}
\colorlet{judgeaccent}{promptaccent}
\colorlet{judgeback}{promptback}
\definecolor{promptslot}{HTML}{A4502A}
\definecolor{methodAWM}{HTML}{F5C796}
\definecolor{methodExpeL}{HTML}{D9700F}
\definecolor{methodRB}{HTML}{E6D29A}
\definecolor{methodACE}{HTML}{B08425}
\definecolor{methodKI}{HTML}{8FD7D7}
\definecolor{methodKT}{HTML}{00A3B0}
\definecolor{methodWT}{HTML}{0A5F68}

\definecolor{pending}{HTML}{7B4FBF}
\colorlet{oursrow}{methodWT!10}
\newsavebox{\fittabularbox}
\newcommand{\fittabular}[4]{\begingroup\setlength{\tabcolsep}{0pt}%
  \sbox{\fittabularbox}{\begin{tabular}{#3}#4\end{tabular}}%
  \setlength{\tabcolsep}{\dimexpr(#1-\wd\fittabularbox)/#2\relax}%
  \ifdim\tabcolsep<1pt\setlength{\tabcolsep}{1pt}\fi
  \begin{tabular}{#3}#4\end{tabular}\endgroup}
\lstdefinestyle{prompttext}{
    language={},
    basicstyle=\fontsize{6.2}{7.8}\selectfont\ttfamily\color{black!85},
    moredelim=[s][\color{promptslot}]{<}{>},
    literate={—}{{\textemdash}}1 {‘}{{\textquoteleft}}1 {’}{{\textquoteright}}1
        {“}{{\textquotedblleft}}1 {”}{{\textquotedblright}}1,
    breaklines=true,
    breakautoindent=false,
    breakindent=0pt,
    columns=fullflexible,
    keepspaces=true,
    showstringspaces=false,
}
\tcbset{
    promptframe/.style 2 args={
        enhanced,
        listing only,
        colback=#2,
        colframe=#1,
        boxrule=0.4pt,
        leftrule=2pt,
        arc=1.5pt,
        coltitle=white,
        fonttitle=\sffamily\bfseries\scriptsize,
        attach boxed title to top left={xshift=8pt, yshift=-\tcboxedtitleheight/2},
        boxed title style={colback=#1, colframe=#1, boxrule=0pt, arc=1pt,
            left=4pt, right=4pt, top=1pt, bottom=1pt},
        listing style=prompttext,
        left=7pt,
        right=7pt,
        top=10pt,
        bottom=6pt,
        before skip=6pt,
        after skip=6pt,
    },
}
\newtcolorbox{defbox}{colback=exampleback, colframe=exampleaccent!60, boxrule=0.4pt, arc=1pt,
  left=5pt, right=5pt, top=3pt, bottom=3pt, fontupper=\small}
\newtcblisting{promptbox}[1][]{promptframe={promptaccent}{promptback}, title={#1}}
\newtcblisting{solverbox}[1][]{promptframe={solveraccent}{solverback}, title={#1}}
\newtcblisting{judgebox}[1][]{promptframe={judgeaccent}{judgeback}, title={#1}}
\definecolor{diffdel}{HTML}{B4452F}
\definecolor{diffadd}{HTML}{2E7D32}
\newtcblisting{solverdiffbox}[1][]{promptframe={solveraccent}{solverback}, title={#1}, listing options={style=prompttext, moredelim={[is][\color{diffdel}]{[-}{-]}}, moredelim={[is][\color{diffadd}]{[+}{+]}}}}
\definecolor{fileaccent}{HTML}{6E5A3C}
\definecolor{fileback}{HTML}{FAF8F4}
\newtcblisting{filebox}[1][]{promptframe={fileaccent}{fileback}, before skip=12pt, after skip=12pt, title={#1}}
\newtcblisting{treebox}[1][]{promptframe={fileaccent}{fileback}, before skip=12pt, after skip=12pt, title={#1}, listing options={style=prompttext, moredelim={[is][\color{promptslot}]{@}{@}}}}
\algrenewcommand{\algorithmiccomment}[1]{\hfill{\color{casegrey}$\triangleright$ #1}}
\definecolor{exampleaccent}{HTML}{5C5F63}
\definecolor{exampleback}{HTML}{F6F6F6}
\newtcblisting{examplebox}[1][]{promptframe={exampleaccent}{exampleback}, before skip=12pt, after skip=12pt, title={#1}}
\definecolor{caseink}{HTML}{20262E}
\definecolor{casegrey}{HTML}{65717A}
\definecolor{casegrid}{HTML}{D9E0E5}
\definecolor{casemark}{HTML}{FFF6D9}
\definecolor{helpink}{HTML}{2E7D4F}
\definecolor{missink}{HTML}{B23A32}
\definecolor{evback}{HTML}{F1F4F8}
\definecolor{evframe}{HTML}{8D9AAA}
\definecolor{insink}{HTML}{38683B}
\definecolor{insback}{HTML}{F7FAF4}
\definecolor{rawink}{HTML}{A33E38}
\definecolor{rawback}{HTML}{FCF1EF}
\definecolor{caselink}{HTML}{2563EB}
\colorlet{wtback}{methodWT!5!white}
\colorlet{kiback}{methodKI!10!white}

\newcommand{\casepar}{\raggedright\hyphenpenalty=10000\relax}

\newcommand{\casereason}[1]{{\itshape\color{black!70}#1}}
\newcommand{\caseact}[1]{{\ttfamily\color{black!88}#1}}
\newcommand{\casetag}[1]{{\sffamily\color{casegrey}#1}}
\newcommand{\markok}{\ensuremath{\checkmark}}
\newcommand{\markno}{\ensuremath{\times}}
\newcommand{\outok}{\textcolor{helpink}{\markok}}
\newcommand{\outno}{\textcolor{missink}{\markno}}
\newlength{\casetagw}

\newtcolorbox{casedecisive}{enhanced, colback=casemark, frame hidden, borderline west={1pt}{0pt}{caseink!55},
  arc=0pt, boxsep=0pt, left=2pt, right=2pt, top=1.5pt, bottom=1.5pt, before skip=1.5pt, after skip=1.5pt,
  grow to left by=3pt, fontupper=\scriptsize\casepar}

\newcommand{\cardfield}[1]{{\scshape\color{black!65}#1}\enspace}
\newcommand{\cardbody}{\normalfont\fontsize{7.2}{8.6}\selectfont\color{black!85}}
\newcommand{\keyph}[1]{{\sethlcolor{casegrid}\hl{#1}}}
\newtcolorbox{insightcard}[1][]{enhanced, colback=insback, colframe=insink!40, boxrule=0.3pt, leftrule=1.2pt,
  arc=1pt, boxsep=0pt, left=3pt, right=3pt, top=2.5pt, bottom=2.5pt, before skip=2pt, after skip=7pt, fontupper=\casepar, #1}
\newcommand{\cardhead}[2]{{\normalfont\fontsize{7.2}{8.6}\selectfont\color{insink}{\ttfamily\bfseries\fontsize{6.8}{8.6}\selectfont #1}\enspace\textbf{#2}}\par\vspace{1pt}}
\newtcolorbox{layercard}{enhanced, colback=evback, colframe=evframe!60, boxrule=0.3pt, leftrule=1.2pt,
  arc=1pt, boxsep=0pt, left=3pt, right=3pt, top=2.5pt, bottom=2.5pt, before skip=4pt, after skip=8pt, fontupper=\casepar}

\newtcolorbox{lanenote}[1][]{enhanced, colback=missink!12!white, frame hidden, borderline west={1pt}{0pt}{missink}, arc=0pt,
  boxsep=0pt, left=3pt, right=2pt, top=1.5pt, bottom=1.5pt, before skip=2pt, after skip=3pt,
  fontupper=\sffamily\fontsize{6.6}{8}\selectfont\color{missink!80!black}\casepar, #1}
\newcommand{\notetag}[1]{{\setlength{\fboxsep}{1pt}\colorbox{missink}{\sffamily\bfseries\fontsize{5.8}{7}\selectfont\color{white}#1}}}
\lstdefinestyle{criteriatext}{style=prompttext, basicstyle=\fontsize{7.6}{9.6}\selectfont\ttfamily\color{black!85}}
\newtcblisting{criteriabox}[1][]{promptframe={judgeaccent}{judgeback}, title={#1}, listing style=criteriatext}
\newtcolorbox{oknote}[1][]{enhanced, colback=helpink!12!white, frame hidden, borderline west={1pt}{0pt}{helpink}, arc=0pt,
  boxsep=0pt, left=3pt, right=2pt, top=1.5pt, bottom=1.5pt, before skip=2pt, after skip=3pt,
  fontupper=\sffamily\fontsize{6.6}{8}\selectfont\color{helpink!80!black}\casepar, #1}
\newtcolorbox{trainnote}[1][]{enhanced, colback=evframe!14!white, frame hidden, borderline west={1pt}{0pt}{evframe}, arc=0pt,
  boxsep=0pt, left=3pt, right=2pt, top=1.5pt, bottom=1.5pt, before skip=2pt, after skip=3pt,
  fontupper=\sffamily\fontsize{6.6}{8}\selectfont\color{evframe!55!black}\casepar, #1}
\newcommand{\oktag}[1]{{\setlength{\fboxsep}{1pt}\colorbox{helpink}{\sffamily\bfseries\fontsize{5.8}{7}\selectfont\color{white}#1}}}

\newlength{\rllabw}
\newlength{\rlgap}
\newcommand{\rlturn}[2]{\par\noindent\makebox[\casetagw][l]{\casetag{#1}}%
  \parbox[t]{\dimexpr\linewidth-\casetagw\relax}{\casepar #2}\par\vspace{2.2pt}\vspace{\rlgap}}
\newcommand{\rllab}[1]{\makebox[\rllabw][l]{\normalfont\scshape\color{black!65}#1}}
\newcommand{\rlr}[1]{\par\noindent\casereason{``#1''}\par}
\newcommand{\rla}[1]{\par\noindent\hangindent=\rllabw\hangafter=1\rllab{Act}\caseact{#1}\par}
\newcommand{\rlo}[1]{\par\noindent\hangindent=\rllabw\hangafter=1\rllab{Obs}{\color{black!60}#1}\par}
\tcbset{titlelane/.style n args={4}{enhanced, colframe=#1, colbacktitle=#2, colback=#3, coltitle=black, title={#4}, boxrule=0.4pt, arc=1.5pt, boxsep=0pt,
  left=4.6pt, right=4.6pt, top=3pt, bottom=3pt, lefttitle=4.6pt, righttitle=4.6pt, toptitle=2.5pt, bottomtitle=2.5pt, before skip=0pt, after skip=0pt,
  fonttitle=\sffamily\bfseries\fontsize{6.4}{7.6}\selectfont, fontupper=\scriptsize\casepar}}
\newtcolorbox{wtlane}[2]{titlelane={methodWT}{methodWT!32!white}{wtback!35!white}{WikiTrace\hfill#1}, equal height group=#2}
\newtcolorbox{kilane}[2]{titlelane={methodKI}{methodKI!55!white}{kiback!35!white}{Insights Only\hfill#1}, equal height group=#2}
\newtcolorbox{rawlane}[2]{titlelane={rawink!45!white}{rawink!25!white}{rawback!35!white}{#1}, equal height group=#2}
\newtcolorbox{withlane}[2]{titlelane={methodKI}{methodKI!55!white}{kiback!35!white}{With memory\hfill#1}, equal height group=#2}
\newtcolorbox{withoutlane}[2]{titlelane={casegrey!50!white}{casegrey!12!white}{casegrey!2!white}{Without memory\hfill#1}, equal height group=#2}
\newcommand{\rltask}[1]{\noindent{\sffamily\tiny\color{casegrey}\textsc{Task}\enspace #1}\par\vspace{3pt}}
\newenvironment{rldecisive}{\begin{casedecisive}\setlength{\rlgap}{0pt}}{\end{casedecisive}\vspace{\rlgap}}
\newcommand{\rlt}[1]{\par\noindent\hangindent=\rllabw\hangafter=1\rllab{Think}\casereason{#1}\par}
\newtcolorbox{memcall}{enhanced, colback=evback, colframe=evframe!70, boxrule=0.3pt, leftrule=1.6pt, arc=1pt, boxsep=0pt,
  left=3pt, right=3pt, top=2pt, bottom=2.5pt, before skip=0.5pt, after skip=2.5pt, fontupper=\scriptsize\casepar}

\colorlet{casekey}{casegrid}

\newtcolorbox{memout}{enhanced, colback=white, colframe=evframe!45, boxrule=0.3pt, arc=0.5pt, boxsep=0pt, left=2.5pt,
  right=2pt, top=1.5pt, bottom=1.5pt, before skip=1.5pt, after skip=0pt,
  fontupper=\ttfamily\fontsize{5.8}{6.9}\selectfont\color{black!80}\raggedright}
\newtcolorbox{dxwhy}[1][]{enhanced, colback=caseink!5, colframe=caseink!70, boxrule=0.5pt, arc=1pt, boxsep=0pt,
  left=4pt, right=4pt, top=2.5pt, bottom=2.5pt, before skip=1pt, after skip=4pt,
  fontupper=\normalfont\fontsize{7.2}{8.8}\selectfont\color{caseink}\casepar, before upper={{\sffamily\bfseries Why}\enspace}, #1}
\newcommand{\dxdescend}{{\setlength{\fboxsep}{1.2pt}\colorbox{caselink}{\sffamily\bfseries\fontsize{6.4}{7.6}\selectfont\color{white}$\downarrow$\,opens the item's source run}}}
\newcommand{\dxdescendrun}{{\setlength{\fboxsep}{1.2pt}\colorbox{caselink}{\sffamily\bfseries\fontsize{6.4}{7.6}\selectfont\color{white}$\downarrow$\,opens a successful run}}}
\newcommand{\dxnote}[2]{\par\vspace{2pt}{\color{#1!40}\hrule height 0.4pt}\vspace{2pt}%
  \noindent{\sffamily\fontsize{6.6}{8}\selectfont\color{#1!85!black}\casepar #2}\par}

\newcommand{\dlabel}[1]{{\setlength{\fboxsep}{0.8pt}\smash{\colorbox{casegrid!60}{\vphantom{Ag}\textcolor{methodWT}{#1}}}}}

\definecolor{linkred}{HTML}{C8553D}
\definecolor{citeblue}{HTML}{2F6FB0}
\hypersetup{
    colorlinks=true,
    urlcolor=citeblue,
    linkcolor=linkred,
    citecolor=citeblue
}

\title{Don't Just Trust Your Memory: Evidence-Linked Memory Reduces Memory-Induced Degradation in LLM Agents}

\author{Yukyeong Lee$^{1*}$ \quad Yumin Choi$^{1*}$ \quad Jinheon Baek$^{1\dagger}$ \quad Sung Ju Hwang$^{1,2\dagger}$ \\
  $^{1}$KAIST \quad $^{2}$DeepAuto.ai \\
  \texttt{\{yukyeong, yuminchoi, jinheon.baek, sungju.hwang\}@kaist.ac.kr}}

\begin{document}
\maketitle
\makeatletter\ifacl@finalcopy\renewcommand{\thefootnote}{\fnsymbol{footnote}}\footnotetext[1]{Equal contribution; $^{\dagger}$Equal advising.}\renewcommand{\thefootnote}{\arabic{footnote}}\fi\makeatother

\begin{abstract}
External memory enables LLM agents to improve from experience by distilling trajectories of past tasks into concise lessons for new tasks.
However, memory can also lead to \emph{memory-induced degradation}, where an agent fails on tasks it would solve without memory, often because it follows a lesson that is incorrect, or misapplies or misreads a valid one.
This is because lessons are inferred from trajectories labeled only with task success or failure, yet are typically provided without the trajectories they came from, leaving the agent no way to check them.
To address this, we introduce \textbf{WikiTrace}, an evidence-linked memory system designed to let agents check each lesson against its original trajectory.
Specifically, motivated by the finding that a trajectory often still holds the evidence needed to catch an erroneous lesson, WikiTrace stores each lesson as a concise insight linked to a summary of why its trajectory succeeded or failed and to the raw trajectory itself.
Then, on a new task, the agent starts from all insights and, whenever one lacks details needed to act, conflicts with another, or comes from a failed task, descends through its links with terminal tools to fill in missing details or correct the insight before acting.
We validate WikiTrace on five agentic benchmarks, showing that it consistently outperforms existing memory approaches and a baseline given the same insights alone, while also reducing memory-induced degradation.
Our code is available at \url{https://github.com/yukyeongleee/wikitrace}.
\end{abstract}

\section{Introduction}
\label{sec:introduction}

Large language model (LLM) agents increasingly solve multi-step tasks by interacting with an environment~\cite{yao2023react}, and external memory allows them to improve from experience without updating their parameters~\cite{sumers2024coala, hu2025memoryagents, luo2026storagetoexperience}. Specifically, memory distills trajectories of past tasks, namely their sequences of actions and observations, into concise lessons that are reused on new tasks, allowing the agent to avoid repeated exploration and to follow procedures specific to its environment. Prior work captures such lessons in diverse forms, including verbal reflections, executable skills, instruction manuals, workflows, and reasoning strategies~\cite{shinn2023reflexion, zhao2024expel, wang2024voyager, chen2024automanual, wang2025awm, ouyang2026reasoningbank, zhang2026ace}.

\begin{figure}[t]
\centering
\includegraphics[width=\columnwidth]{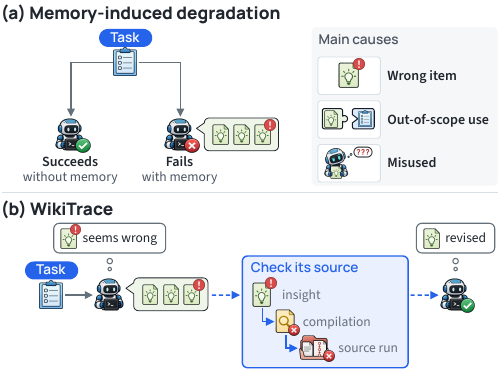}
\caption{\textbf{Memory-induced degradation and WikiTrace.} (a) The agent solves a task without memory but fails with memory that holds a wrong insight. (b) With WikiTrace, the agent follows the wrong insight's links to its source run, revises it, and solves the task.}
\label{fig:teaser}
\vspace{-0.1in}
\end{figure}

However, memory does not always help and can even lead to \emph{memory-induced degradation}, where an agent fails on tasks it would solve without memory~\cite{zhang2026faultymemories, xiong2025memorymanagement, hu2026continualmemory} (Figure~\ref{fig:teaser}a). Reliable experience reuse therefore requires solving more tasks while preserving those the agent already solves, but average accuracy cannot tell the two apart. We thus measure them separately, as \emph{Gain}, the rate at which memory turns failures into successes, and \emph{Regression}, the rate at which it turns successes into failures. As shown in Figure~\ref{fig:gain-regression-ci}, existing memory methods exhibit Regression comparable to, or even higher than, that of rerunning the agent without memory.

To understand this degradation, we analyze tasks that the agent consistently solves without memory but fails with it, and find that 52\% of the failures trace to a memory item the agent followed (Figure~\ref{fig:regression-causes}), most often a wrong one. Wrong items arise because a trajectory comes with only a single label, success or failure, which does not reveal which of its many actions caused that outcome. Any method that learns lessons from such a trajectory therefore has to guess which actions matter and under which conditions, and each wrong guess yields a lesson that misleads the agent on new tasks. Yet, existing approaches provide lessons without the trajectories they came from, leaving the agent no way to check them, even though the evidence needed to catch an erroneous lesson often remains in that trajectory.

To this end, we present \textbf{WikiTrace}, an evidence-linked memory system that keeps each lesson together with a path back to its original trajectory (Figures~\ref{fig:teaser}b and~\ref{fig:overview}), combining the strengths of both: a lesson concisely tells the agent what to do, whereas the trajectory faithfully records what actually happened but is too long to reread for every decision. Specifically, WikiTrace organizes its memory into three layers linked from top to bottom: each lesson is stored as a concise \emph{insight}, which links down to a \emph{compilation} that summarizes why its trajectory succeeded or failed, and further down to the \emph{raw trajectory} itself. Then, on a new task, the agent starts from all insights and, whenever an insight lacks details needed to act, conflicts with another, or comes from a failed trajectory, descends through its links with terminal tools to recover missing details or revise the insight before acting.

We validate WikiTrace on five agentic benchmarks comprising eight suites, spanning simulated text worlds~\cite{shridhar2021alfworld, wang2022scienceworld, dagan2025plancraft} and tool-use environments~\cite{yao2025taubench, barres2026tau2bench, trivedi2024appworld}. WikiTrace achieves the best performance on every suite, improving over the agent without memory by 10.7 points on average and outperforming a baseline given the same insights alone by 3.2 points, while reducing memory-induced degradation (Regression from 6.5\% to 4.5\%). Further analyses show that the agent consults the linked evidence more often on harder tasks, where its advantage is largest, and gains most when the evidence changes its plan. These gains hold under another backbone.

\section{Related Work}
\label{sec:related-work}

\subsection{Experience-Driven Procedural Memory}

Language agents increasingly convert past execution experience into procedural memory that can guide future decisions.
Prior methods retain verbal feedback or successful trajectories~\cite{shinn2023reflexion,zhao2024expel}, accumulate executable skills~\cite{wang2024voyager}, or distill experience into reusable guidelines, manuals, workflows, modular procedures, and multigranular scripts~\cite{fu2024autoguide,chen2024automanual,wang2025awm,han2026legomem,fang2026memp}.
Continual approaches further update reasoning memories and playbooks with new successful and failed experience~\cite{zhou2025memento,ouyang2026reasoningbank,zhang2026ace,cao2026reme}.
These approaches develop mechanisms for distilling, selecting, and updating experience for reuse.
WikiTrace focuses on reassessing the resulting guidance during task execution, linking each lesson to its source evidence so that the agent can examine its basis and applicability to the current task.

\begin{figure*}[t]
\centering
\includegraphics[width=\textwidth]{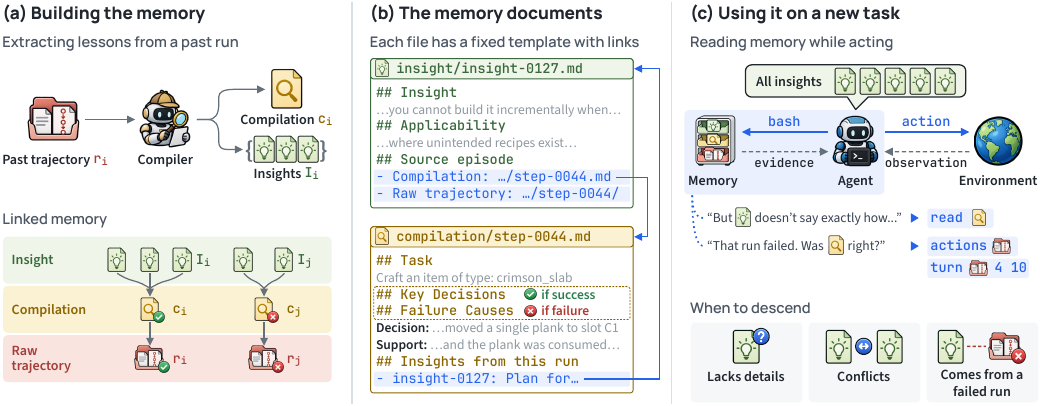}
\caption{\textbf{Overview of WikiTrace.} (a) A compiler turns each past run into a compilation, a diagnosis of why the run succeeded or failed, and insights, each a concise lesson. All layers stay linked. (b) Each layer's documents follow a fixed template. (c) The agent starts from every insight and descends through the links under the conditions shown.}
\label{fig:overview}
\end{figure*}

\subsection{Reliable Reuse of Unverified Memory}
\label{sec:related-reliable}

Reusing past experience does not always improve an agent's performance.
Agents ignore or misinterpret relevant condensed experience~\cite{zhao2026faithful}, encounter retrieval competition and negative transfer during continual learning~\cite{hu2026continualmemory}, and exhibit amplified unsafe behavior when reusing execution-oriented experience collected from benign tasks~\cite{zhao2026safety}.
These findings motivate evaluating the failures introduced by memory alongside the tasks it helps solve.
We explicitly separate Gain from Regression to highlight memory-induced degradation that aggregate accuracy obscures.
Research on memory admission and revision addresses reliability by controlling which experiences and lessons are retained.
PREPING validates synthetic practice trajectories before storage~\cite{choi2026preping}, SkillOpt admits skill updates through held-out evaluation~\cite{yang2026skillopt}, and ReMe updates or removes memories using subsequent utility~\cite{cao2026reme}.
WikiTrace complements these approaches by focusing on how stored lessons inform decisions during task execution.
This perspective treats reliable memory reuse as a question of both which lessons to retain and whether those lessons are justified in the situations where they are applied.

\subsection{Memory Structure and Access}

Agent memory design concerns both how past experience is organized and how agents access it during task execution.
Hierarchical approaches organize experience at multiple levels of detail for compression, planning, and retrieval~\cite{hu2025hiagent,zhang2025gmemory,fang2026memp,yang2025coarsetofine,gao2025cdmem,hu2026compassmem}, while HORMA connects filesystem summaries to raw interaction history and learns to select a minimal sufficient context~\cite{hsu2026horma}.
Beyond memory organization, active-access methods determine when and what to retrieve during reasoning~\cite{zhang2026expseek,zhao2026expweaver,cai2026proactagent}.
Direct Corpus Interaction (DCI) provides a general approach to corpus access through terminal tools~\cite{li2026dciagent}.
WikiTrace brings these two concerns together around the reassessment of procedural lessons: its structure preserves the connection between a lesson and its source execution, and its access mechanism lets the agent consult that execution when deciding how to apply the lesson.

\section{WikiTrace}
\label{sec:method}

\noindent We first define the problem setting and examine memory-induced degradation to motivate WikiTrace. We then describe its memory construction and test-time evidence access (Figure~\ref{fig:overview}).

\subsection{Problem Setting}
\label{sec:method-scope}

We consider an agent that reuses a fixed corpus of past runs collected without memory,
\[
\mathcal{D}_{\mathrm{train}}=\{(x_i,r_i,s_i)\}_{i=1}^{N}.
\]
Each run consists of a task $x_i$, a trajectory $r_i$ recording the agent's actions and observations, and a binary outcome $s_i\in\{0,1\}$ indicating task success or failure. No gold solutions or expert explanations are available, so a run's outcome does not exactly tell us why it succeeded or failed. Thus, an insight drawn from it is not always valid.

We construct memory $\mathcal{M}$ from this corpus before any test task runs, with no additional run to validate an individual memory item. Once constructed, the memory remains fixed and read-only for an agent solving new tasks. Our objective is to improve task performance through experience reuse while limiting memory-induced degradation on tasks the agent solves without memory.

\subsection{Understanding Memory-Induced Degradation}
\label{sec:analysis-regression}

\begin{figure*}[t]
\newcommand{\capthree}{\textbf{Failure categories in memory-induced degradation.} LLM-judged counts of failures on tasks solved in all three trials without memory.}%
\newcommand{\capfour}{\textbf{A wrong lesson induces degradation.} A rule in memory says an ALFWorld lamp task ends once the object sits beside the lit lamp, which its source run contradicts. Following it, the agent fails a task it solves without memory (Figure~\ref{fig:wrong-items}).}%
\sbox0{\begin{minipage}[t]{\columnwidth}\vspace{0pt}\input{figures/fig04-regression-example}\end{minipage}}%
\sbox1{\begin{minipage}{\columnwidth}\caption*{Figure~3: \capthree}\end{minipage}}%
\sbox2{\begin{minipage}{\columnwidth}\caption*{Figure~4: \capfour}\end{minipage}}%
\begin{minipage}[t]{\columnwidth}\vspace{0pt}\centering
\includegraphics[height=\dimexpr\ht0+\dp0+\ht2+\dp2-\ht1-\dp1\relax,width=\linewidth,keepaspectratio]{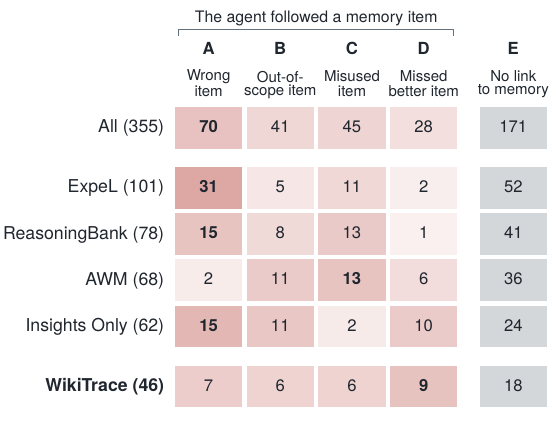}
\caption{\capthree}
\label{fig:regression-causes}
\end{minipage}\hfill
\begin{minipage}[t]{\columnwidth}\vspace{0pt}\usebox0
\caption{\capfour}
\label{fig:regression-example}
\end{minipage}
\end{figure*}

To understand how memory breaks tasks the agent can already solve, we examine tasks that the agent solves in all three repeated runs without memory but fails with memory. An LLM judge, \texttt{gpt-5.6-sol} at medium reasoning effort, compares each failed run with a no-memory run of the same task, locates the first turn where the two runs depart, and classifies the cause into one of five kinds (Figure~\ref{fig:regression-causes}; Appendix~\ref{sec:appendix-case-regression}).

In 52\% of these cases, \textit{the failure traces to a memory item the agent followed} (A to D). Most often the item itself was wrong (A, 20\%). Otherwise, a valid item was applied outside its scope (B), misused by the agent (C), or followed while memory held a better one (D).\footnote{A cause is often implicit, and an agent acts differently across runs even without memory. Many cases therefore show no link to any memory item (E).} Checking the original run a lesson was drawn from can reveal a wrong or out-of-scope item. Figure~\ref{fig:regression-example} shows an example: a wrong rule from ExpeL~\cite{zhao2024expel} fails a task the agent solves without memory, although the run behind the rule contradicts it.

These observations motivate preserving a path from each insight back to its source run. With this path, the agent can recover the observations an insight omits and check whether the insight is supported and whether it applies to the current task before following it. WikiTrace realizes this through source-linked memory (Section~\ref{sec:method-construction}) and test-time evidence access (Section~\ref{sec:method-access}).

\begin{table*}[t]
\centering
\caption{
\textbf{Main results.}
Mean over three trials, $\pm$ the standard deviation across trials in small type.
Avg.\ averages the eight suites.
Best results are bolded and second-best results are underlined.
}
\label{tab:main-results}

\begingroup
\small
\setlength{\tabcolsep}{2.5pt}
\renewcommand{\arraystretch}{1.15}
\newcommand{\pmcell}[1]{{\color{gray}\fontsize{7}{8}\selectfont $\pm$#1}}
\fittabular{\textwidth}{20}{l*{9}{c}}{%
\toprule
& \multicolumn{2}{c}{\textbf{ALFWorld}} & \multirow{2}{*}[-\dimexpr(\aboverulesep+\belowrulesep+\cmidrulewidth)/2\relax]{\textbf{ScienceWorld}} & \multicolumn{2}{c}{\textbf{PlanCraft}} & \multirow{2}{*}[-\dimexpr(\aboverulesep+\belowrulesep+\cmidrulewidth)/2\relax]{\textbf{\makecell{$\tau^3$-bench\\Retail}}} & \multicolumn{2}{c}{\textbf{AppWorld}} & \multirow{2}{*}[-\dimexpr(\aboverulesep+\belowrulesep+\cmidrulewidth)/2\relax]{\textbf{Avg.}} \\
\cmidrule(lr){2-3}
\cmidrule(lr){5-6}
\cmidrule(lr){8-9}
\textbf{Method} & \textbf{Seen} & \textbf{Unseen} &  & \textbf{Seen} & \textbf{Unseen} &  & \textbf{Normal} & \textbf{Challenge} &  \\
\midrule
No Memory & 94.4{\pmcell{2.0}} & 94.3{\pmcell{1.9}} & 83.7{\pmcell{0.8}} & 72.3{\pmcell{1.9}} & 56.3{\pmcell{1.2}} & \underline{86.8{\pmcell{1.9}}} & 75.4{\pmcell{3.4}} & 72.7{\pmcell{2.2}} & 79.5{\pmcell{1.6}} \\
\noalign{\vskip 2.5pt}
\hdashline[2pt/2pt]
\noalign{\vskip 2.5pt}
AWM & 94.7{\pmcell{1.2}} & 97.3{\pmcell{0.9}} & 88.9{\pmcell{4.7}} & 76.3{\pmcell{3.7}} & 57.3{\pmcell{2.1}} & 85.5{\pmcell{2.1}} & 79.3{\pmcell{0.7}} & 73.4{\pmcell{1.7}} & 81.6{\pmcell{0.3}} \\
ReasoningBank & 92.1{\pmcell{2.4}} & 92.8{\pmcell{0.7}} & 85.0{\pmcell{1.3}} & 84.7{\pmcell{2.6}} & 61.3{\pmcell{3.4}} & 86.4{\pmcell{1.6}} & 80.8{\pmcell{1.5}} & 70.7{\pmcell{3.0}} & 81.7{\pmcell{0.5}} \\
ExpeL & 94.4{\pmcell{0.9}} & \underline{98.0{\pmcell{0.4}}} & 91.9{\pmcell{1.6}} & 74.0{\pmcell{8.5}} & 51.0{\pmcell{3.6}} & 82.5{\pmcell{5.1}} & 80.5{\pmcell{5.7}} & 70.0{\pmcell{3.7}} & 80.3{\pmcell{1.6}} \\
\noalign{\vskip 2.5pt}
\hdashline[2pt/2pt]
\noalign{\vskip 2.5pt}
Insights Only & \underline{96.7{\pmcell{0.5}}} & 97.3{\pmcell{1.3}} & \underline{94.9{\pmcell{0.6}}} & \underline{88.7{\pmcell{6.1}}} & \underline{73.0{\pmcell{10.7}}} & 85.1{\pmcell{2.7}} & \underline{86.8{\pmcell{0.4}}} & \underline{73.7{\pmcell{1.4}}} & \underline{87.0{\pmcell{2.5}}} \\
\rowcolor{oursrow}\textbf{WikiTrace (ours)} & \textbf{98.7{\pmcell{0.5}}} & \textbf{98.8{\pmcell{0.7}}} & \textbf{96.1{\pmcell{1.4}}} & \textbf{93.3{\pmcell{3.4}}} & \textbf{81.3{\pmcell{8.5}}} & \textbf{88.2{\pmcell{3.9}}} & \textbf{89.5{\pmcell{1.5}}} & \textbf{75.8{\pmcell{2.2}}} & \textbf{90.2{\pmcell{1.6}}} \\
\bottomrule
}
\endgroup
\end{table*}

\subsection{Constructing Source-Linked Memory}
\label{sec:method-construction}

WikiTrace constructs memory $\mathcal{M}=(R,C,I,E)$, consisting of raw trajectories $R$, compilations $C$, insights $I$, and source links $E$. Each run is processed independently so that every extracted lesson remains connected to the experience it came from.

\paragraph{Compilation and insight extraction}
A compiler reads a task $x_i$, the agent's trajectory $r_i$ on that task, and a binary signal $s_i$ of its success or failure, producing a compilation $c_i$ and a set of insights $I_i$:
\[
(c_i,I_i)=\mathrm{Compiler}(x_i,r_i,s_i).
\]
The compilation summarizes the run's key decisions and supporting observations, using separate templates for successful and failed runs~\cite{ouyang2026reasoningbank, yang2026skillopt}. For successful runs, it describes decisions associated with success; for failed runs, it records plausible explanations for failure. These accounts remain interpretations of the run rather than verified explanations of its outcome. The compiler also extracts at most three insights per run. Each insight is a separate document that says what to do, when it applies, which run it came from, and whether that run succeeded. We generate insights independently for each run, so each insight has a clear source, and we never update or delete them once written. The insight pool may therefore contain redundant or conflicting insights, and deciding which one applies to the current task is left to the agent.

\paragraph{Source preservation and linking}
WikiTrace preserves each raw trajectory unchanged: the task metadata, each turn's reasoning, action and observation, and whether the run succeeded. Each insight links to both its source compilation and raw trajectory. Thus, when reassessing an insight, the test-time agent can see which run it came from and how that run was interpreted to derive it. To help the agent navigate, each layer also has an index. Appendix~\ref{sec:appendix-memory-system} provides the corpus layout, document templates, and construction algorithm.

\subsection{Revisiting Source Evidence at Test Time}
\label{sec:method-access}

WikiTrace lets the agent start from the insights and descend to their source runs when an insight may not hold or does not say enough to act on now.

\paragraph{Initial input}
The agent first receives all insights, each stating what to do, when it applies, and the id of its source run. The message then lists the three past runs whose tasks are most similar to the current one, with the ids and titles of the insights drawn from each. These runs give the agent starting points for its search.

\paragraph{Descent to source runs}
To let the agent examine an insight before following it, the prompt asks it to \emph{descend} from the insight to its source run when the insight lacks the details needed to act on it, comes from a failed run, or disagrees with another insight. It asks for another descent whenever an action fails. The source-linked memory is mounted as read-only files, and the prompt encourages the agent to search them with \texttt{read} and \texttt{bash} for the fact that settles the question, such as a command format. Figure~\ref{fig:prompt-memory-wikitrace} in Appendix~\ref{sec:appendix-prompts} gives the full prompt.
\section{Experiments}
\label{sec:experiments}

\begin{figure}[t]
\centering
\includegraphics[width=\columnwidth]{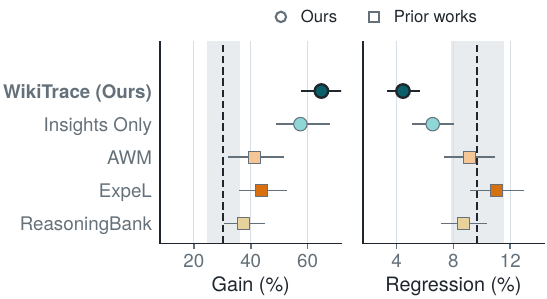}
\caption{\textbf{Gain and Regression of WikiTrace and memory-based baselines}, averaged over eight suites with 95\% intervals.}
\label{fig:gain-regression-ci}
\vspace{-0.1in}
\end{figure}

\subsection{Experimental Setup}
\label{sec:experiments-protocol}

\paragraph{Benchmarks}
We evaluate on eight suites across five benchmarks. Three are simulated text worlds, where the agent issues one text command at a time: ALFWorld~\cite{shridhar2021alfworld} (household tasks), ScienceWorld~\cite{wang2022scienceworld} (science experiments) and PlanCraft~\cite{dagan2025plancraft} (Minecraft crafting). Two are tool-use environments: AppWorld~\cite{trivedi2024appworld} poses everyday tasks through application APIs, and $\tau^3$-bench Retail~\cite{yao2025taubench,barres2026tau2bench} combines conversation with a simulated customer and tool calls over an order database. AppWorld includes Test-Normal and Test-Challenge; the latter requires unseen applications. Appendix~\ref{sec:appendix-suite-construction} details suite construction and splits.

\paragraph{Baselines}
\textbf{No Memory} is the same agent with nothing added to its prompt or files.
\textbf{AWM}~\cite{wang2025awm} abstracts successful trajectories into reusable workflows. \textbf{ExpeL}~\cite{zhao2024expel} keeps a list of free-form rules and reviews the whole list against each successful trajectory to add, edit or remove rules. \textbf{ReasoningBank}~\cite{ouyang2026reasoningbank} distills successful and failed trajectories into reasoning strategies and retrieves them by task similarity. We adapt their official implementations to our offline protocol (Appendix~\ref{sec:appendix-compared-methods}). \textbf{Insights Only} gives the same insights as WikiTrace but mounts no source compilations or raw trajectories and omits the prompt to descend, so the two differ only in access to source runs.

\paragraph{Protocol}
We run three trials per suite and method. In each trial, every method builds its memory once from the same No Memory runs of the training tasks and is tested on the same test tasks. No method may make additional runs to validate its memory. All methods use the same agent: a Pi terminal agent~\cite{pi_coding_agent} given each benchmark's ReAct instruction. The agent and memory builders use \texttt{deepseek-v4-flash}\footnote{The 2026-04-23 release.} with high reasoning effort. We report each benchmark's original metric, averaged over the three trials. See Appendix~\ref{sec:appendix-agent-harness} for implementation details.

\subsection{Main Results}
\label{sec:experiments-results}

Table~\ref{tab:main-results} shows that WikiTrace achieves the highest performance on all eight suites, averaging 90.2 compared with 79.5 for No Memory. It also exceeds Insights Only by 3.2 points on average, demonstrating the benefit of source evidence and the descent prompt beyond the same insights alone. Notably, WikiTrace is the only memory method that matches or exceeds No Memory on every suite, suggesting that revisiting source evidence helps mitigate memory-induced degradation.

\begin{figure}[t]
\centering
\includegraphics[width=\columnwidth]{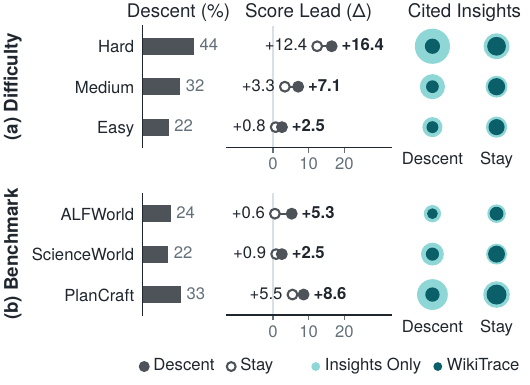}
\caption{\textbf{WikiTrace against Insights Only by descent} in the text-world benchmarks. Score Lead $\Delta$ is WikiTrace minus Insights Only, and circle radius gives the insights reported as used. Difficulty: No Memory solves the task in none (Hard), one or two (Medium) or all three (Easy) trials.}
\label{fig:descent-lead}
\end{figure}

\subsection{Gain and Regression}
\label{sec:experiments-gain-regression}

Aggregate performance does not distinguish improvements on previously failed tasks from failures on previously solved tasks. We therefore report Gain and Regression separately.

\begin{defbox}
\textbf{Gain and Regression.} For task $i$, let $B_i$ denote the majority No Memory outcome across three trials and $M_{i,t}$ the outcome with memory in trial $t$. Gain measures success on tasks No Memory fails, while Regression measures failure on tasks No Memory solves:
\[
\begin{aligned}
\mathrm{Gain} &= \Pr_{i,t}(M_{i,t}=1 \mid B_i=0),\\
\mathrm{Regression} &= \Pr_{i,t}(M_{i,t}=0 \mid B_i=1).
\end{aligned}
\]
\end{defbox}

Figure~\ref{fig:gain-regression-ci} reports both rates averaged across suites, with 95\% bootstrap confidence intervals over tasks (Appendix~\ref{sec:appendix-ci}). To account for variation across runs, we evaluate three additional No Memory trials against the same reference outcomes $B_i$. The dashed line shows this rerun baseline, measuring outcome changes without memory.

WikiTrace achieves a Regression of 4.5\%, compared with 6.5\% for Insights Only and 9.7\% for the rerun baseline. Its Regression is lower than Insights Only on seven of the eight suites, and it is the only method whose Regression confidence interval lies entirely below the rerun band. This reduction accompanies higher Gain: WikiTrace exceeds the rerun baseline by 35 percentage points, compared with 27 for Insights Only and 7 to 14 for existing memory methods. Because WikiTrace and Insights Only differ only in access to source runs, the higher Gain and lower Regression are attributable to that access. Relative to Insights Only, wrong or out-of-scope items cause half as many of WikiTrace's regressions (Figure~\ref{fig:regression-causes}). Appendix~\ref{sec:appendix-gain-regression-suite} provides results for individual suites.

\section{Analysis}
\label{sec:analysis}

\subsection{When Does Descent Help?}
\label{sec:analysis-descent}

Insights Only gives the agent the same insights, so comparing it with WikiTrace on the same task and trial shows what access to source runs adds. We split WikiTrace's test runs in the three text-world benchmarks into Descent runs, which open a compilation or raw trajectory, and Stay runs, which do not (Figure~\ref{fig:descent-lead}). WikiTrace descends more often on harder tasks, in 22\% of runs on Easy tasks and 44\% on Hard ones. Descent runs lead Insights Only by 1.6 to 4.7 points more than Stay runs at every difficulty and on every benchmark. The gap between Descent and Stay grows with difficulty, from 1.8 points on Easy tasks to 3.7 on Medium and 4.1 on Hard ones. These results suggest that source runs help more on harder tasks. The agent chooses when to descend, however, so Descent and Stay runs differ in more than the descent.

Stay runs also lead Insights Only, by 0.6 to 12.4 points. In Stay runs, WikiTrace reports 19 to 28\% fewer insights as used,\footnote{After each run, the agent names the insights it applied.} which suggests more selective use of memory. The reduction is larger in Descent runs, 38 to 57\%.

\begin{figure}[!t]
\centering
\includegraphics[width=\columnwidth]{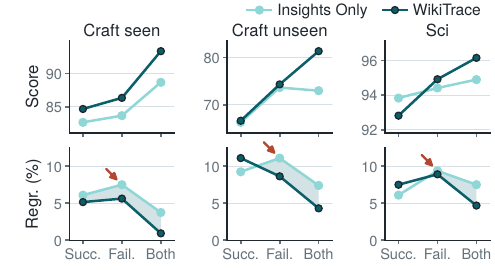}
\caption{\textbf{Score and Regression by memory subset.} Memory is built from successful runs (Succ.), failed runs (Fail.) or both. Arrows mark Insights Only's higher Regression with failed runs.}
\label{fig:outcome-subset}
\end{figure}

\subsection{Why and How Does WikiTrace Descend?}
\label{sec:analysis-when}
\label{sec:analysis-traversal}

Section~\ref{sec:analysis-descent} shows when descent helps, but not why the agent descends or what the opened run changes. To find out, an LLM judge reads the agent's reasoning around each descent in the three text-world benchmarks and labels three things: its \textit{trigger}, what prompted the descent; its \textit{target}, what the agent looked for in the opened run; and its \textit{effect}, what the run contributed to the agent's next actions (Figures~\ref{fig:descent-before} and~\ref{fig:descent-after}). The judge, \texttt{gpt-5.6-sol} at medium reasoning effort, agrees with a second judge, \texttt{GLM-5.3}, at Cohen's $\kappa$ of 0.77 on trigger and 0.63 on target. Effect is harder to judge, so its label is the majority of three judges (Appendix~\ref{sec:appendix-descent-judge}).

\begin{figure}[t]
\input{figures/fig08-descent-before}
\vspace{8pt}
\input{figures/fig09-descent-after}
\end{figure}%

\paragraph{Why it descends and what it seeks}
WikiTrace's advantage over Insights Only concentrates on descents made after a problem. Descents that follow a blocked attempt, an observation that blocked the agent's plan, or a doubt about an insight make up 36\% of descents and lead Insights Only by 10.4 points, against 3.7 points for descents made with nothing gone wrong (Figure~\ref{fig:descent-before}). The largest advantage, 17.3 points, comes when a blocked attempt sends the agent after a specific fact in past runs, which Insights Only has no way to look up. Routine checks are the most common descent (38\%), and 90\% of them read a similar run's procedure, with a small advantage of 3.4 points.

\paragraph{How it uses source evidence}

The opened run changes the agent's plan in 43\% of descending runs, by revising an insight (24\%), adding a detail an insight leaves out (14\%), or supplying a fact no insight covers (4\%). These runs lead Insights Only by 8.3 points, against 4.4 points when the opened run only confirms the plan or goes unused (Figure~\ref{fig:descent-after}). Figure~\ref{fig:case-study} illustrates such a revision, with additional cases in Appendix~\ref{sec:appendix-descent-cases}.

\subsection{Can Agents Reliably Learn from Failed Experiences?}
\label{sec:analysis-outcome}
\begin{table}[t]
\centering
\small
\setlength{\tabcolsep}{4pt}
\newcommand{\pmcell}[1]{{\color{gray}\fontsize{7}{8}\selectfont $\pm$#1}}
\begin{tabular}{@{}lcccc@{}}
\toprule
& \multicolumn{2}{c}{\textbf{Craft}} & \multirow{2}{*}[-\dimexpr(\aboverulesep+\belowrulesep+\cmidrulewidth)/2\relax]{\textbf{Sci}} & \multirow{2}{*}[-\dimexpr(\aboverulesep+\belowrulesep+\cmidrulewidth)/2\relax]{\textbf{Avg.}} \\
\cmidrule(lr){2-3}
\textbf{Method} & \textbf{Seen} & \textbf{Unseen} & & \\
\midrule
NM & 68.7{\pmcell{3.3}} & 55.3{\pmcell{2.1}} & 89.6{\pmcell{1.5}} & 71.2{\pmcell{0.3}} \\
\midrule
AWM & 66.0{\pmcell{2.8}} & 47.0{\pmcell{2.9}} & 91.7{\pmcell{1.0}} & 68.2{\pmcell{0.5}} \\
ReasoningBank & 74.7{\pmcell{4.2}} & 53.3{\pmcell{5.2}} & 90.9{\pmcell{1.9}} & 73.0{\pmcell{2.3}} \\
ExpeL & 79.0{\pmcell{12.8}} & 67.0{\pmcell{14.1}} & 88.6{\pmcell{1.4}} & 78.2{\pmcell{8.5}} \\
\midrule
Insights Only & 92.3{\pmcell{2.1}} & 87.7{\pmcell{3.3}} & 92.6{\pmcell{0.8}} & 90.9{\pmcell{1.6}} \\
\rowcolor{oursrow}\textbf{WT} & \textbf{95.7}{\pmcell{1.7}} & \textbf{89.0}{\pmcell{1.6}} & \textbf{93.3}{\pmcell{0.8}} & \textbf{92.7}{\pmcell{0.4}} \\
\bottomrule
\end{tabular}
\caption{\textbf{Results with \texttt{GLM-5.3-flash}}. Mean over three trials. NM is No Memory and WT is WikiTrace.}
\label{tab:other-model}
\end{table}

\begin{figure}[t]
\centering
\begin{minipage}[c]{1.7in}
\includegraphics{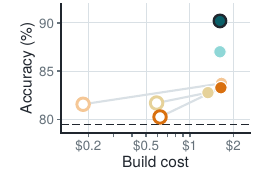}\end{minipage}\hfill
\begin{minipage}[c]{\dimexpr\columnwidth-1.71in}\raggedleft\scriptsize
\setlength{\tabcolsep}{2pt}
\begin{tabular}{@{}c@{\hspace{3pt}}lc@{}}
\toprule
\multicolumn{2}{@{}l}{\textbf{Method}} & \textbf{Avg.} \\
\midrule
\tikz[baseline=-0.6ex]{\path[use as bounding box] (-3.000pt,-3.000pt) rectangle (3.000pt,3.000pt);\draw[draw={rgb,255:red,32;green,38;blue,46},line width=0.600pt,dash pattern=on 2.400pt off 1.200pt] (-3.000pt,0)--(3.000pt,0);} & No Memory & 79.5 \\
\midrule
\tikz[baseline=-0.6ex]{\path[use as bounding box] (-3.000pt,-3.000pt) rectangle (3.000pt,3.000pt);\filldraw[fill={rgb,255:red,245;green,199;blue,150},draw={rgb,255:red,255;green,255;blue,255},line width=0.333pt] (0,0) circle (2.000pt);} & AWM ($\times$9) & 83.8 \\
\tikz[baseline=-0.6ex]{\path[use as bounding box] (-3.000pt,-3.000pt) rectangle (3.000pt,3.000pt);\filldraw[fill={rgb,255:red,230;green,210;blue,154},draw={rgb,255:red,255;green,255;blue,255},line width=0.333pt] (0,0) circle (2.000pt);} & ReasoningBank ($\times$3) & 82.8 \\
\tikz[baseline=-0.6ex]{\path[use as bounding box] (-3.000pt,-3.000pt) rectangle (3.000pt,3.000pt);\filldraw[fill={rgb,255:red,217;green,112;blue,15},draw={rgb,255:red,255;green,255;blue,255},line width=0.333pt] (0,0) circle (2.000pt);} & ExpeL ($\times$3) & 83.3 \\
\midrule
\tikz[baseline=-0.6ex]{\path[use as bounding box] (-3.000pt,-3.000pt) rectangle (3.000pt,3.000pt);\filldraw[fill={rgb,255:red,143;green,215;blue,215},draw={rgb,255:red,255;green,255;blue,255},line width=0.333pt] (0,0) circle (2.000pt);} & Insights Only & 87.0 \\
\rowcolor{oursrow}\tikz[baseline=-0.6ex]{\path[use as bounding box] (-3.000pt,-3.000pt) rectangle (3.000pt,3.000pt);\filldraw[fill={rgb,255:red,10;green,95;blue,104},draw={rgb,255:red,32;green,38;blue,46},line width=0.733pt] (0,0) circle (2.000pt);} & \textbf{WikiTrace} & \textbf{90.2} \\
\bottomrule
\end{tabular}

\end{minipage}
\caption{\textbf{Accuracy against build cost}, averaged over the eight suites. A hollow mark is a default build and a filled mark the build run $N$ times. The dashed line shows No Memory.}
\label{fig:cost-accuracy}
\end{figure}

Failed runs contain useful experience, but the lessons drawn from them depend on how the failure is interpreted. To examine their value for memory, we compare WikiTrace and Insights Only using insights from successful runs, failed runs, or both on PlanCraft and ScienceWorld. Figure~\ref{fig:outcome-subset} shows that both methods perform better with failed runs alone than with successful runs alone, and perform best with both. Failed runs therefore add guidance beyond what successful runs give.

However, this benefit accompanies a greater risk of regression when the extracted insights are provided alone. Insights Only has its highest Regression with failed runs alone, reaching 9.3\% on PlanCraft and 9.4\% on ScienceWorld. With the same insights, WikiTrace reduces these rates to 7.1\% and 8.9\%. These findings support keeping failed runs with their source evidence, so that agents can revisit the basis of a lesson before reusing it.

\subsection{How Cost-Effective Is WikiTrace?}
\label{sec:analysis-cost}
\begin{figure*}[t]
\definecolor{caseleft}{HTML}{F3F5F7}
\colorlet{csmistake}{missink!20!white}
\newcommand{\cslabel}[1]{{\sffamily\bfseries\fontsize{6.2}{7.4}\selectfont\color{casegrey}#1}}
\newcommand{\cstitle}[2]{\par\noindent\tikz[baseline=(n.base)]\node[circle, fill=caseink, text=white, inner sep=0pt, minimum size=8.5pt, font=\sffamily\bfseries\fontsize{6.4}{7.6}\selectfont] (n) {#1};\enspace{\sffamily\bfseries\fontsize{7.4}{9}\selectfont\color{caseink}#2}\par\vspace{3pt}}
\newcommand{\csview}[2][64pt]{\includegraphics[page=#2,width=#1]{figures/exported/fig11-case-study-frames.pdf}}
\newcommand{\csframe}[2]{\parbox[t]{64pt}{\vspace{0pt}\csview{#1}\par\vspace{1pt}\makebox[64pt][c]{\cslabel{#2}}}}
\newcommand{\csframes}[1]{\par\noindent\makebox[\linewidth][l]{#1}\par}
\newcommand{\mc}[1]{{\mdseries\ttfamily\detokenize{#1}}\,\raisebox{-.15em}{\includegraphics[height=1.05em]{figures/exported/items/#1}}}
\newcommand{\csblue}[1]{{\sffamily\bfseries\fontsize{6.6}{8}\selectfont\color{caselink}#1}}
\newcommand{\csmistakeword}[1]{{\setlength{\fboxsep}{0.8pt}\colorbox{csmistake}{\textbf{#1}}}}
\begin{tcolorbox}[enhanced, nobeforeafter, box align=top, width=96pt, colback=caseleft, frame hidden, arc=1.5pt, boxsep=0pt, left=3pt, right=3pt, top=3pt, bottom=3pt, equal height group=csmain]
\cslabel{Task}\enspace{\normalfont\fontsize{6.4}{7.6}\selectfont\color{caseink}craft \mc{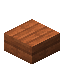}}\par\vspace{10pt}
\cslabel{Solution}\par\vspace{1pt}
\csview[\linewidth]{1}\par\vspace{10pt}
\cslabel{Agent's Common Mistake}\par\vspace{1pt}
{\normalfont\fontsize{6.8}{8.2}\selectfont\color{caseink}\casepar\csmistakeword{Auto-craft}: After one item is placed, the output shows a product.\par\vspace{2pt}
\noindent\makebox[\linewidth][c]{\csview{2}}\par\vspace{3pt}
Agents think this preview is an already crafted item and try to remove it.\par\vspace{2pt}
\noindent\makebox[\linewidth][c]{\csview{3}}\par\vspace{3pt}
Taking it uses up the item, so the agent can run out of materials and fail.\par}\vspace{10pt}
\cslabel{Memory (Partial View)}\par\vspace{1pt}
\includegraphics[width=\linewidth]{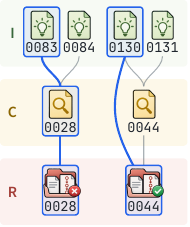}
\end{tcolorbox}\hfill
\begin{tcolorbox}[nobeforeafter, box align=top, width=354pt, blank, equal height group=csmain]
\cardbody\casepar
\cstitle{1}{A wrong insight}
\begin{minipage}[t]{166pt}\vspace{0pt}
\begin{insightcard}[equal height group=csA, before skip=0pt, after skip=0pt]\cardhead{insight-0083}{failed run\enspace\notetag{Wrong}}{\cardbody If placing a single item already crafts something, a recipe that needs several of it may be impossible. Consider giving up.}\end{insightcard}
\end{minipage}\hfill
\begin{minipage}[t]{181pt}\vspace{0pt}
\begin{dxwhy}[equal height group=csA, before skip=0pt, after skip=0pt]\mc{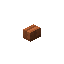} in the output is only a preview of what the grid matches so far. It was not \csmistakeword{auto-crafted}: \mc{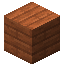} stays on the grid. Ignore the preview and finish the recipe.\end{dxwhy}
\end{minipage}\par\vspace{24pt}
\cstitle{2}{What each method did next}
\begin{minipage}[t]{140.5pt}\vspace{0pt}
\begin{kilane}{\markno}{csB}
\rlt{``the plank was consumed to make the button.''}\vspace{2pt}
\csframes{\csframe{3}{Takes \mc{acacia_button}}\hspace{2.5pt}\parbox[t]{64pt}{\vspace{0pt}\begin{tikzpicture}\node[inner sep=0pt, outer sep=0pt] (view) {\csview{12}};\fill[white, opacity=0.88] (view.south west) rectangle (view.north east);\node[align=center, text=missink, font=\sffamily\bfseries\fontsize{7.4}{9}\selectfont] at (view.center) {Declares\\impossible};\end{tikzpicture}}}
\dxnote{caseink}{Taking it uses up 1 \mc{acacia_planks}.}
\end{kilane}
\end{minipage}\hfill
\begin{minipage}[t]{207pt}\vspace{0pt}
\begin{wtlane}{\markok}{csB}
\rlt{``But wait, let me also look at the successful run \ldots''}\vspace{2pt}
\csframes{\csframe{4}{2 \mc{acacia_planks}}\hspace{2.5pt}\csframe{5}{3 \mc{acacia_planks}}\hspace{2.5pt}\csframe{6}{Takes \mc{acacia_slab}}}
\dxnote{caseink}{\mc{acacia_planks} are used up only when it takes \mc{acacia_slab}.}
\end{wtlane}
\end{minipage}\par\vspace{24pt}
\cstitle{3}{How WikiTrace revised the insight}
\begin{insightcard}[before skip=0pt, after skip=3pt]\cardhead{insight-0130}{successful run}{\cardbody Until the recipe is complete, the output shows only an intermediate item. Place all required items.}\end{insightcard}
\rlt{``This lesson {\setlength{\fboxsep}{0.8pt}\colorbox{caselink!15!white}{\textbf{conflicts with insight-0083}}}. Let's check the evidence of both.''}\vspace{3pt}
\noindent{\setlength{\fboxsep}{1.2pt}\colorbox{caselink}{\sffamily\bfseries\fontsize{6.4}{7.6}\selectfont\color{white}$\downarrow$\,Descent}}\enspace\csblue{opens the source run of each insight}\par\vspace{3pt}
\begin{minipage}[t]{140.5pt}\vspace{0pt}
\begin{rawlane}{Source of insight-0083 (step-0028)\hfill\markno}{csC}
\rltask{craft \mc{acacia_slab}, different inventory}
\csframes{\csframe{7}{1 \mc{acacia_planks}}\hspace{2.5pt}\csframe{8}{Takes it back}}
\dxnote{caseink}{Repeats this 9~times until its moves run out.}
\end{rawlane}
\end{minipage}\hfill
\begin{minipage}[t]{207pt}\vspace{0pt}
\begin{rawlane}{Source of insight-0130 (step-0044)\hfill\markok}{csC}
\rltask{craft \mc{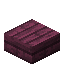}}
\csframes{\csframe{9}{1 \mc{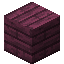}}\hspace{2.5pt}\csframe{10}{2 \mc{crimson_planks}}\hspace{2.5pt}\csframe{11}{3 \mc{crimson_planks}}}
\dxnote{caseink}{Keeps adding \mc{crimson_planks}. The output changes until \mc{crimson_slab} appears.}
\end{rawlane}
\end{minipage}\par\vspace{9pt}
\rlt{``The button/pressure plate outputs are temporary and get replaced when the full slab pattern is formed. So I can safely place my \mc{acacia_planks} one at a time!''}\par\vspace{2pt}\noindent\csblue{\texttt{insight-0083} is wrong: it is OK to keep adding \mc{acacia_planks}.}
\end{tcolorbox}
\caption{\textbf{Qualitative example: revising a wrong insight (PlanCraft).} Insight-0083 leads agents into the auto-craft mistake. Insights Only follows it and fails, but WikiTrace revises it through descent and succeeds.}
\label{fig:case-study}
\end{figure*}

To examine whether WikiTrace's advantage comes from greater memory construction cost, we increase existing methods' construction budgets by running their builders over the same training trajectories nine times for AWM and three times for ExpeL and ReasoningBank. Figure~\ref{fig:cost-accuracy} shows that WikiTrace retains the highest average performance even with these larger baseline budgets. At test time, WikiTrace costs 1.17 times as much per task as Insights Only, averaged across suites, while improving average performance by 3.2 points. These results suggest that WikiTrace improves performance with modest additional test-time cost.

\subsection{Does WikiTrace Generalize?}
\label{sec:analysis-models}

WikiTrace also achieves the highest performance on all three evaluated suites with \texttt{GLM-5.3-flash} (Table~\ref{tab:other-model}). We evaluate all methods on PlanCraft and ScienceWorld at medium reasoning effort, using training executions generated by the same model to construct memory. WikiTrace averages 92.7, exceeding Insights Only by 1.8 percentage points and No Memory by 21.5. Each existing memory method falls below No Memory on at least one suite. This pattern mirrors the DeepSeek results, supporting the benefit of source-linked memory across the two models.

\subsection{Resolving Conflicting Insights}
\label{sec:case-study}

Figure~\ref{fig:case-study} illustrates how WikiTrace resolves conflicting insights by revisiting source evidence. In PlanCraft, an insight from a failed run incorrectly rules out a valid procedure that another recommends. The successful run's observations show that the procedure works. WikiTrace revises the insight and succeeds, whereas Insights Only, given identical insights, follows the wrong guidance and fails. This case shows how raw trajectories enable agents to reassess interpretations embedded in insights and compilations.

\section{Conclusion}
\label{sec:conclusion}

Lessons extracted from past executions are not validated by their outcomes alone, and reusing them without scrutiny can cause memory-induced degradation. We introduced WikiTrace, which links insights to source compilations and raw trajectories so that agents can recover missing details and reassess their guidance. Across five benchmarks, WikiTrace achieves the highest performance on every suite while increasing Gain and reducing Regression compared with injecting the same insights alone. These findings suggest that agent memory should preserve not only reusable lessons, but also the evidence needed to reassess them.

\section*{Limitations}
Memory is built from binary trajectory outcomes, so a compilation's diagnosis of a failure, and any insight drawn from it, may be wrong. A source link lets the agent check an insight against its run but does not verify the insight. WikiTrace places every insight in the first message so that the agent never misses one, and the prompt therefore grows with the memory. A memory too large to place in full would need the agent to find the insights it needs, a retrieval problem this work leaves open. The causes of degradation (Section~\ref{sec:analysis-regression}) and the descent labels (Section~\ref{sec:analysis-when}) come from an LLM judge that reads the agent's reasoning, so they describe what that reasoning states, not what produced the outcome.

\clearpage\twocolumn
\bibliography{reference}

\clearpage
\appendix
\raggedbottom

\twocolumn[{\section*{Appendix}\begin{center}
\captionof{table}{
\textbf{Benchmark suites.}
One training set per benchmark serves all of its test suites.
}
\label{tab:env-suites}

\small
\renewcommand{\arraystretch}{1.15}
\begin{tabular}{@{}llrr>{\raggedright\arraybackslash}p{0.42\textwidth}l@{}}
\toprule
\textbf{Benchmark} & \textbf{Suite} & \textbf{Train} & \textbf{Test} & \textbf{What the test holds out} & \textbf{Metric} \\
\midrule
\multirow[t]{2}{*}{ALFWorld}
& \texttt{valid-seen}     & 50 & 101 & Games in training rooms. & Success rate \\
& \texttt{valid-unseen}   & 50 & 134 & Games in rooms absent from training. & Success rate \\
\midrule
ScienceWorld
& \texttt{scienceworld} & 50 & 100 & Descriptions absent from training, same task types. & Mean score \\
\midrule
\multirow[t]{2}{*}{PlanCraft}
& \texttt{plancraft-seen}   & 50 & 100 & Examples with training target items. & Success rate \\
& \texttt{plancraft-unseen} & 50 & 100 & Examples with target items absent from training. & Success rate \\
\midrule
\multirow[t]{2}{*}{AppWorld}
& \texttt{test-normal}    & 54 & 111 & Tasks over training apps. & TGC \\
& \texttt{test-challenge} & 54 &  99 & Tasks needing apps absent from training. & TGC \\
\midrule
$\tau^3$-bench
& \texttt{retail} & 38 & 76 & New requests, mostly from training customers. & Pass rate \\
\bottomrule
\end{tabular}
\end{center}
\vspace{1em}}]

\section{Experimental Details}
\label{sec:appendix}

\subsection{Benchmark Suite Construction}
\label{sec:appendix-suite-construction}

Table~\ref{tab:env-suites} lists the suites. Every draw is seeded.

\paragraph{ALFWorld}
Training holds 50 games of the training split, eight or nine per task type, in rooms that also appear in valid\_seen. \texttt{valid-seen} is the full set of 101 valid\_seen games filtered to rooms visited by the 50 training games, and \texttt{valid-unseen} uses all 134 valid\_unseen games, set in four rooms absent from training.

\paragraph{ScienceWorld}
We use ten of the 30 task types, boil, melt, freeze, change-the-state-of-matter-of, use-thermometer, measure-melting-point-known-substance, measure-melting-point-unknown-substance, test-conductivity, test-conductivity-of-unknown-substances, and chemistry-mix. Each meets three conditions.
\begin{itemize}[leftmargin=1.2em,itemsep=0pt,parsep=0pt,topsep=2pt]
  \item Its variations pose distinct tasks.
  \item Its training and test descriptions are largely disjoint.
  \item Its reference solutions fit the simulator's 100-step limit.
\end{itemize}
Training holds five variations per type from the training split, and test holds 100 variations from the test split whose description does not occur in training.

\paragraph{PlanCraft}
We use solvable examples only, weighted toward medium and hard ones. Training holds 50 examples of the training split with distinct target items that also occur in the test split. \texttt{plancraft-seen} holds 100 test examples whose target is a training target, and \texttt{plancraft-unseen} 100 whose target occurs in no solvable training example.

\paragraph{AppWorld}
A scenario groups three tasks and is solved only when all three pass, so every suite keeps scenarios whole. We use difficulty 2 and 3 only. Training holds all 18 such training scenarios (54 tasks), \texttt{test-normal} all 37 of the Test-Normal split (111 tasks), and \texttt{test-challenge} a draw of 33 of the 65 difficulty-3 scenarios of the Test-Challenge split (99 tasks). TGC counts tasks and SGC scenarios.

\paragraph{$\tau^3$-bench (Retail)}
We set aside the authors' splits and divide the 114 tasks into 38 training and 76 test tasks, grouped by customer, so every customer with several tasks appears on both sides. The simulated user is \texttt{gpt-4.1}, and the judge of conversation assertions is \texttt{gpt-5.4-mini} at medium reasoning effort.

\subsection{Agent Harness}
\label{sec:appendix-agent-harness}

\begin{table*}[t]
\centering
\caption{
\textbf{Agent harness per benchmark.}
The instruction, exemplars, and step budget follow the listed ReAct baseline. A step is one environment tool call, except in $\tau^3$-bench, which counts agent, user, and tool turns as its authors do.
}
\label{tab:harness}

\small
\renewcommand{\arraystretch}{1.15}
\begin{tabular}{@{}lllrr@{}}
\toprule
\textbf{Benchmark} & \textbf{ReAct baseline} & \textbf{Tool} & \textbf{Max steps} & \textbf{Max Pi turns} \\
\midrule
ALFWorld       & ReAct loop~\cite{yao2023react} & \texttt{act} & 49 & 150 \\
ScienceWorld   & SwiftSage's ReAct baseline~\cite{lin2023swiftsage} & \texttt{act} & 100 & 400 \\
PlanCraft      & The authors' prompt~\cite{dagan2025plancraft} & \texttt{take\_action} & 30 & 150 \\
AppWorld       & AppWorld's ReAct code agent~\cite{trivedi2024appworld} & \texttt{exec} & 50 & 150 \\
$\tau^3$-bench & $\tau^2$-bench's LLM agent~\cite{barres2026tau2bench} & \texttt{take\_action} & 200 & 150 \\
\bottomrule
\end{tabular}
\end{table*}

Table~\ref{tab:harness} lists each benchmark's harness. An episode ends when the task is done, the step budget runs out, or the agent replies without a tool call. The Pi turn cap is only a guard, since a turn may hold memory reads that are not steps.

\paragraph{Tool \texttt{act(command)}}
Sends one text command to the game and returns its observation. We add a \textit{done} command that ends the episode, so the agent can stop early.

\paragraph{Tool \texttt{take\_action(action, argument)}}
Joins the action and its argument with a colon for the benchmark's own parser. PlanCraft's actions are \textit{move} and \textit{smelt} (source slot, target slot, quantity), \textit{search} (a recipe, seeded), and \textit{impossible}, which ends the episode. We drop the authors' \textit{think}, since the agent has its own reasoning channel. In $\tau^3$-bench an action is one of the 16 retail tools, or \textit{respond} with a message to the simulated user. An episode also ends on the user's stop, transfer, or out-of-scope token, or after 10 tool errors.

\paragraph{Tool \texttt{exec(code)}}
Runs Python in the task's shell, which holds the benchmark's API bindings and keeps state across calls, and returns the output.

\subsection{Compared Methods: Details}
\label{sec:appendix-compared-methods}

\begin{table}[t]
\centering
\caption{
\textbf{Compared methods.}
Formatted Rules are rules with required fields.
WikiTrace links each to the compilation and raw trajectory it came from.
Access is inject (I), retrieve (R) or traverse (T).
}
\label{tab:compared-methods}

\begingroup
\fontsize{8}{10}\selectfont
\renewcommand{\arraystretch}{1.0}
\setlength{\aboverulesep}{0.2ex}\setlength{\belowrulesep}{0.5ex}
\newcommand{\no}{{\color{gray!60}--}}
\fittabular{0.9\columnwidth}{10}{l c c l c}{%
\toprule
& \multicolumn{2}{c}{\textbf{Learns from}} & & \\
\cmidrule(lr){2-3}
\textbf{Method} & \textbf{Succ.} & \textbf{Fail.} & \textbf{Memory} & \textbf{Access} \\
\midrule
AWM                 & \checkmark & \no        & Procedures & I \\
ExpeL               & \checkmark & \no        & Rules & I \\
ReasoningBank & \checkmark & \checkmark & Formatted Rules & R \\
\midrule
Insights Only & \checkmark & \checkmark & Formatted Rules & I \\
\rowcolor{oursrow} WikiTrace (ours) & \checkmark & \checkmark & + Linked Evidence & I + T \\
\bottomrule
}
\endgroup
\end{table}

We run the official implementations of the existing memory methods with their own induction prompts and retrieval~\cite{wang2025awm,zhao2024expel,ouyang2026reasoningbank}, changed only to fit our protocol (Section~\ref{sec:experiments-protocol}).
\begin{itemize}[leftmargin=1.2em,itemsep=0pt,parsep=0pt,topsep=2pt]
  \item \textbf{Offline.} AWM runs in its offline mode, which induces workflows once from training examples. ReasoningBank updates memory after every test task, and we instead build it once. Every method thus builds memory once from the shared no-memory training trajectories and freezes it.
  \item \textbf{True outcome.} Each builder receives the binary outcome, in place of an LLM judge (AWM, ReasoningBank) or the environment reward (ExpeL).
  \item \textbf{Single pass.} Each training trajectory is read once. ExpeL thus has no failed and successful attempts at one task to compare, which its Reflexion~\cite{shinn2023reflexion} retries would provide, so it draws insights from successes alone, and we omit its exemplar retrieval.
\end{itemize}

\clearpage
\section{Memory System Details}
\label{sec:appendix-memory-system}

\subsection{Corpus and Memory Layout}
\label{sec:appendix-corpus-layout}

\begin{figure*}[t]
\centering
\begin{examplebox}[raw/step-0000/solver.full.jsonl, excerpt]
{"type": "message", "role": "system", "content": "Interact with a household to solve a task. Here ...", "turn": 0}
{"type": "message", "role": "user", "content": "Here is the task.\nYou are in the middle of a roo ...", "turn": 1}
{"turn": 2, "type": "function_call", "role": "assistant", "name": "act", "arguments": {"command": "go to cabinet 1"}, "reasoning": "Let me start by exploring the room to find a kettle. I'll ch ...", "result": "You arrive at cabinet 1. The cabinet 1 is closed.", "is_error": false}
{"turn": 3, "type": "function_call", "role": "assistant", "name": "act", "arguments": {"command": "open cabinet 1"}, "reasoning": "Let me open cabinet 1 to see what's inside.", "result": "You open the cabinet 1. The cabinet 1 is open. In it, you se ...", "is_error": false}
<turns 4 to 20: seventeen more act calls of the same form>
{"turn": 21, "type": "function_call", "role": "assistant", "name": "act", "arguments": {"command": "move kettle 1 to shelf 1"}, "reasoning": "I can put the kettle on shelf 1.", "result": "You move the kettle 1 to the shelf 1.\n\nEpisode over", "is_error": false}
{"turn": 22, "type": "episode_over"}
{"type": "message", "role": "assistant", "content": "", "reasoning": "", "turn": 24}
{"type": "answer", "role": "assistant", "model_requested": null, "model_returned": "deepseek/deepseek-v4-flash", "parsed": null, "usage": {"prompt_tokens": 52444, "completion_tokens": 1172, "...": "..."}, "read_set": [], "turn": 25}
\end{examplebox}
\caption{\textbf{Excerpt of \texttt{raw/step-0000/solver.full.jsonl}, the turn log of one ALFWorld training run.} Long strings are cut short.}
\label{fig:raw-jsonl}
\end{figure*}

\begin{algorithm}[t]
\caption{\small\textbf{\textsc{WikiTrace}: Memory Construction}}
\label{alg:wikitrace-construction}
\footnotesize
\begin{algorithmic}[1]
\Require Fixed corpus $\mathcal{D}_{\mathrm{train}}=\{(x_i,r_i,s_i)\}_{i=1}^{N}$
\Ensure Frozen evidence-linked memory corpus $\mathcal{M}$
\State $R \gets \{r_i\}_{i=1}^{N}$, $C \gets \varnothing$, $I \gets \varnothing$, $E \gets \varnothing$
\For{$i=1,\ldots,N$}
    \State $(c_i, I_i) \gets \Call{Compiler}{x_i,r_i,s_i}$
    \State $C \gets C \cup \{c_i\}$, $I \gets I \cup I_i$
    \State $E \gets E \cup \{(u, i) \mid u \in I_i\}$
\EndFor
\State Build $\textproc{Idx}(L)$ for $L \in \{R,C,I\}$ \Comment{document titles}
\State $\mathcal{M} \gets \Call{Freeze}{R,C,I,E}$ \Comment{read-only}
\State \Return $\mathcal{M}$
\end{algorithmic}
\end{algorithm}

WikiTrace mounts its corpus read-only at \texttt{/work/corpus}. Each index is generated by code from the layer it lists, and the corpus has no root index.

\begin{treebox}[WikiTrace corpus layout]
/work/corpus/
|-- corpus.json               @build stamp@
|-- raw/
|   |-- _index.md             @every run and its outcome@
|   `-- step-NNNN/
|       |-- metadata.md       @task, outcome, step count@
|       `-- solver.full.jsonl @the agent's turn log@
|-- compilation/
|   |-- _index.md             @every compilation by title@
|   `-- step-NNNN.md          @one per run@
`-- insight/
    |-- _index.md             @insight titles by task type@
    `-- insight-NNNN.md       @one per candidate insight@
\end{treebox}

A raw run keeps the record the agent left, unchanged. For one ALFWorld training run, which succeeded in 20 \texttt{act} calls, the box below shows its \texttt{metadata.md} and Figure~\ref{fig:raw-jsonl} an excerpt of its \texttt{solver.full.jsonl}. The turn log holds the system and user messages, one line per environment call with the model's reasoning and the observation, and a closing usage record.

\begin{examplebox}[raw/step-0000/metadata.md]
# alfworld_0000_pick_and_place_simple-Kettle-None-Shelf-1

- task_type: pick_and_place_simple
- instruction: Your task is to: put a kettle in shelf.
- initial_observation: You are in the middle of a room. Looking quickly around you, you see a cabinet 9, a cabinet 8, a cabinet 7, a cabinet 6, a cabinet 5, a cabinet 4, a cabinet 3, a cabinet 2, a cabinet 1, a coffeemachine 1, a countertop 3, a countertop 2, a countertop 1, a drawer 9, a drawer 8, a drawer 7, a drawer 6, a drawer 5, a drawer 4, a drawer 3, a drawer 2, a drawer 1, a fridge 1, a garbagecan 1, a microwave 1, a shelf 3, a shelf 2, a shelf 1, a sinkbasin 1, a stoveburner 4, a stoveburner 3, a stoveburner 2, a stoveburner 1, and a toaster 1.
- success: True
- steps: 20 act calls
\end{examplebox}

\subsection{Evidence-Linked Document Formats}
\label{sec:appendix-document-formats}

The compiler's structured output is rendered as markdown with a frontmatter block. In the mounted corpus, a compilation lists links to the insight documents made from its candidate insights in place of the insights themselves.

\begin{filebox}[Success compilation]
---
type: trajectory-compilation
title: [success] <title of the case>
description: <one-line description>
step_id: step-NNNN
outcome: success
---

# Compilation (success)

## Task
<the run's task fields>

## Key Decisions
### Decision 1
**Decision**: <what the agent chose to do>
**Support**: <what showed it was necessary>

## Insights from this run
- insight-NNNN: <title> (/work/corpus/insight/insight-NNNN.md)
\end{filebox}

\begin{filebox}[Failure compilation]
---
type: trajectory-compilation
title: [failure] <title of the case>
description: <one-line description>
step_id: step-NNNN
outcome: failure
---

# Compilation (failure)

## Task
<the run's task fields>

## Failure Causes
### Cause 1
**Decision**: <what the agent did, or did not do>
**Support**: <what showed it caused the failure>

## Insights from this run
- insight-NNNN: <title> (/work/corpus/insight/insight-NNNN.md)
\end{filebox}

\begin{filebox}[Insight document]
---
type: insight
id: insight-NNNN
title: <insight title>
task_type: <task type of the source run>
source_step: step-NNNN
---

## Insight
<what to do>

## Applicability
<when it applies, and when it does not>

## Source episode
step-NNNN (<success|failure>)
- Task: <the source run's instruction>
- Compilation: /work/corpus/compilation/step-NNNN.md
- Raw trajectory: /work/corpus/raw/step-NNNN/
\end{filebox}

On AppWorld an insight document carries \texttt{apps}, the apps its source run called, in place of \texttt{task\_type}.

\subsection{Construction Algorithm}
\label{sec:appendix-construction-algorithm}

WikiTrace builds its memory in one pass over a fixed corpus of no-memory trajectories, with one compiler call per trajectory (Algorithm~\ref{alg:wikitrace-construction}). Each candidate insight becomes its own document, tagged with its source run and that run's outcome and task type. Insights are neither merged nor filtered, so $|I|=\sum_i |I_i|$.

\clearpage
\section{Gain and Regression with Confidence Intervals}
\label{sec:appendix-gain-regression-suite}

\subsection{Confidence Intervals}
\label{sec:appendix-ci}

\begin{figure*}[t]
\centering
\includegraphics[width=\textwidth]{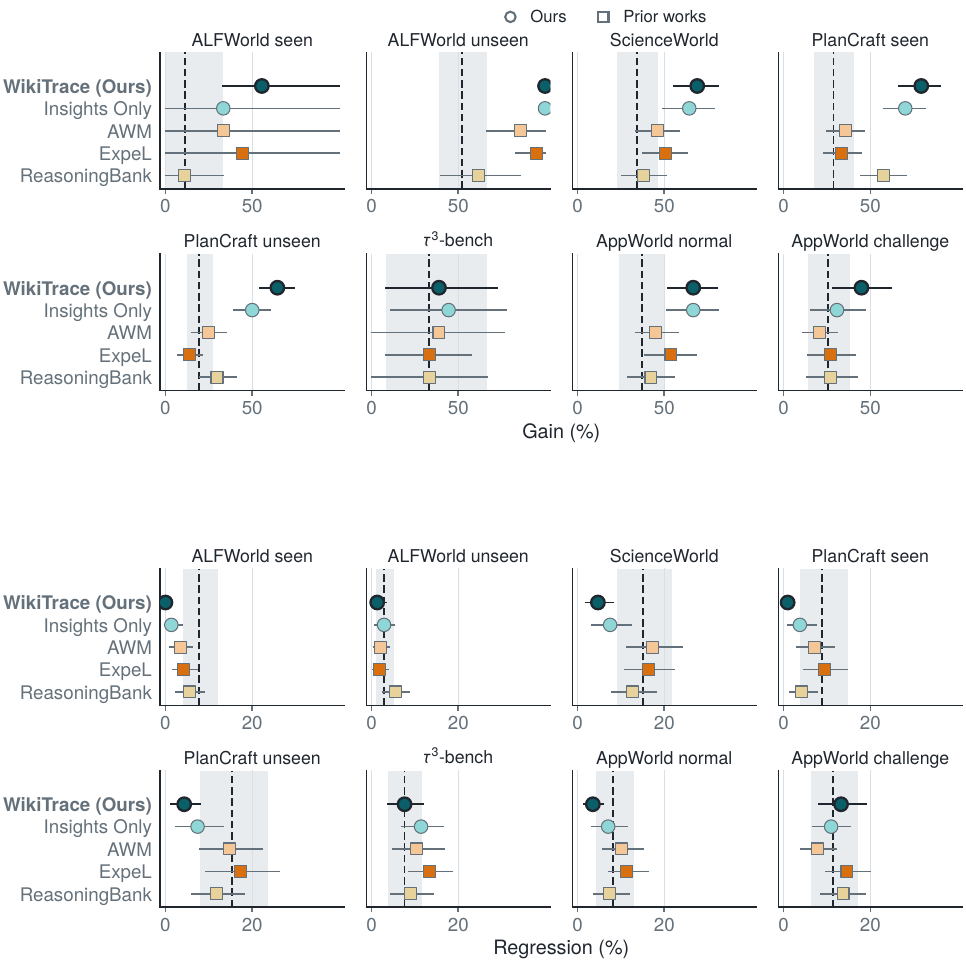}
\caption{\textbf{Gain and Regression per suite}, with 95\% intervals. The dashed line and band denote rerunning without memory.}
\label{fig:gain-regression-ci-bench}
\end{figure*}

Table~\ref{tab:gain-regression} counts the tasks that each method moved per benchmark, the numerators and denominators behind the rates. Figure~\ref{fig:gain-regression-ci-bench} draws the rates of Section~\ref{sec:experiments-gain-regression} for each suite with 95\% confidence intervals and the No Memory floor.

\paragraph{Intervals.} We resample tasks with replacement within each suite 10,000 times. A resampled task keeps its three No Memory trials and its three trials with each method, and every rate is recomputed on the same resample, so the methods stay paired. The suite average is taken within each resample, and its interval spans the 2.5th to 97.5th percentiles of these averages.\footnote{Resamples with no eligible ALFWorld-seen task for Gain (4.8\%) are dropped.}

\paragraph{No Memory floor.} $B_i$ is the majority of three No Memory trials, so a fresh run without memory can disagree with it by chance. We ran three further No Memory trials and score them against the same $B_i$ exactly as a method: the Regression floor is the share of their trials that fail a task with $B_i=1$, and the Gain floor the share that solve a task with $B_i=0$. Its interval comes from the same task resamples, in which each task keeps all six No Memory trials. A method's excess is its rate minus the floor, with the interval of that difference over the resamples. A positive excess is added Gain, or, for Regression, added harm.

\begin{table*}[t]
\centering
\scriptsize
\setlength{\tabcolsep}{3.2pt}
\newcommand{\ratecell}[1]{{\color{gray}\fontsize{5.6}{6}\selectfont (#1\%)}}
\caption{\textbf{Gain and Regression counts per benchmark.} Tasks moved on one trial, averaged over trials and summed over suites, with the rate over the eligible tasks (first row) in small type. Bold is best and underlined second best.}
\label{tab:gain-regression}
\begin{tabular}{@{}lcccccccccc@{}}
\toprule
& \multicolumn{2}{c}{\textbf{ALFWorld}} & \multicolumn{2}{c}{\textbf{ScienceWorld}} & \multicolumn{2}{c}{\textbf{PlanCraft}} & \multicolumn{2}{c}{\textbf{$\tau^3$-bench}} & \multicolumn{2}{c}{\textbf{AppWorld}} \\
\cmidrule(lr){2-3}\cmidrule(lr){4-5}\cmidrule(lr){6-7}\cmidrule(lr){8-9}\cmidrule(lr){10-11}
\textbf{Method} & Gain $\uparrow$ & Reg. $\downarrow$ & Gain $\uparrow$ & Reg. $\downarrow$ & Gain $\uparrow$ & Reg. $\downarrow$ & Gain $\uparrow$ & Reg. $\downarrow$ & Gain $\uparrow$ & Reg. $\downarrow$ \\
\midrule
{\color{gray}Eligible tasks} & {\color{gray}10} & {\color{gray}225} & {\color{gray}29} & {\color{gray}71} & {\color{gray}75} & {\color{gray}125} & {\color{gray}6} & {\color{gray}70} & {\color{gray}52} & {\color{gray}158} \\
\midrule
AWM & 7.0 \ratecell{70} & 6.0 \ratecell{3} & 13.3 \ratecell{46} & 12.3 \ratecell{17} & 21.7 \ratecell{29} & 13.0 \ratecell{10} & \underline{2.3} \ratecell{39} & 7.3 \ratecell{10} & 17.0 \ratecell{33} & 14.3 \ratecell{9} \\
ReasoningBank & 4.7 \ratecell{47} & 12.3 \ratecell{5} & 11.0 \ratecell{38} & 9.0 \ratecell{13} & 30.3 \ratecell{40} & 9.3 \ratecell{7} & 2.0 \ratecell{33} & \underline{6.3} \ratecell{9} & 18.0 \ratecell{35} & 16.3 \ratecell{10} \\
ExpeL & \underline{8.0} \ratecell{80} & 6.3 \ratecell{3} & 14.7 \ratecell{51} & 11.7 \ratecell{16} & 16.0 \ratecell{21} & 16.0 \ratecell{13} & 2.0 \ratecell{33} & 9.3 \ratecell{13} & 21.0 \ratecell{40} & 20.3 \ratecell{13} \\
\midrule
Insights Only & \underline{8.0} \ratecell{80} & \underline{5.0} \ratecell{2} & \underline{18.7} \ratecell{64} & \underline{5.3} \ratecell{8} & \underline{43.3} \ratecell{58} & \underline{6.7} \ratecell{5} & \textbf{2.7} \ratecell{44} & 8.0 \ratecell{11} & \underline{25.3} \ratecell{49} & \underline{14.0} \ratecell{9} \\
\rowcolor{oursrow}WikiTrace & \textbf{8.7} \ratecell{87} & \textbf{1.7} \ratecell{1} & \textbf{20.0} \ratecell{69} & \textbf{3.3} \ratecell{5} & \textbf{52.7} \ratecell{70} & \textbf{3.0} \ratecell{2} & \underline{2.3} \ratecell{39} & \textbf{5.3} \ratecell{8} & \textbf{29.0} \ratecell{56} & \textbf{12.7} \ratecell{8} \\
\bottomrule
\end{tabular}
\end{table*}

\clearpage
\section{Causes of Memory-Induced Degradation}
\label{sec:appendix-regression}
\label{sec:appendix-case-regression}

A judge, \texttt{gpt-5.6-sol} at medium reasoning effort, labels every strong degradation case of Section~\ref{sec:analysis-regression} with the system prompt of Figure~\ref{fig:regression-criteria}. The user message holds the case's files. These are the case metadata, the run with memory, the run without memory on the same task and trial, the memory as the agent saw it, and the evaluation of both runs.

The judge finds the turn where the run with memory first departs from the run without memory and gives the case one of five causes. Figures~\ref{fig:wrong-items} to~\ref{fig:regression-labels-e} show one case of each cause.
\begin{itemize}[nosep, leftmargin=2.2em]
\item[(A)] \textbf{Wrong item.} The item states something false about the environment.
\item[(B)] \textbf{Out-of-scope item.} A sound item is applied outside its scope.
\item[(C)] \textbf{Misused item.} A sound item in scope is misread or stretched.
\item[(D)] \textbf{Missed better item.} A sound item is followed while memory holds a better one.
\item[(E)] \textbf{No link to memory.} No item explains the departure.
\end{itemize}

\begin{figure*}[p]
\centering
\begin{judgebox}[System prompt of the regression judge]
You investigate one case of memory-induced regression and return one verdict as JSON.

# The case

An agent ran one test task three times without memory and succeeded every time. It then ran the same task with memory and failed. Your job is to decide whether the memory caused that failure, and if so how.

The user message holds the case's files, each after a line "=== <file name> ===". Those lines are not part of any file.

- `case.json` gives the benchmark, suite, trial, task id and how the memory reached the agent.
- `memory_run.md` is the failed run with memory. A header gives the task and how the episode ended, then each turn gives the agent's reasoning, its action and a clipped observation.
- `nomem.md` is a successful run without memory on the same task and trial, in the same format.
- `memory.txt` is the memory exactly as the agent saw it, up to about 110,000 characters. The memory sat in the agent's first user message, before the task. For some methods a list of the most similar past runs followed the task in the same message, and `memory.txt` includes it after the memory.
- `eval_memory.json` and `eval_nomem.json` record how each episode was scored.

The run without memory is an independent run of the same task, not a replay. Two runs can differ by chance, so a difference counts as caused by memory only under the evidence rule.

# Items

An item is any unit of memory the agent saw: an insight, a workflow, a rule or a memory item in `memory.txt`, a past run listed there, or a past run or its summary that the agent opened during the episode. Cite an item by its id, for example `Workflow 3`, `Memory Item 2`, `insight-0039`, `4.`, `[si-00012]` or `step-0016`.

# Evidence rule

Memory caused the divergence only if at least one of these holds.

- e1. The agent's reasoning in the memory run refers to an item, by id, by title or by a close paraphrase.
- e2. The agent's actions follow steps an item prescribes and depart from what the run without memory did at that point.

An item that is merely about the same topic as the task is not evidence.

# Decision procedure

Apply the steps in order and stop at the first that matches.

1. Find the divergence. It is the first turn where the memory run departs from the run without memory in a way that leads to the failure.
2. If neither e1 nor e2 holds for that divergence, the label is E.
3. If the item the agent followed states something false about this environment, or prescribes a rule that hurts in the very tasks it claims to cover, the label is A.
4. If the item the agent followed is sound, but memory also held another item that fits this task better and, followed instead, would have avoided the divergence, the label is D.
5. If the item is sound where it came from but its stated or implied scope does not cover this task, the label is B.
6. If its scope covers this task and the agent misread it, applied it to steps it does not cover, or let it push aside a requirement the task states, the label is C.

When several items are involved, label by the one that most directly produced the divergence and list the others too.

# Labels

The examples come from ScienceWorld. The labels apply to every benchmark in the same way.

- A, a wrong or harmful item. This includes an item distilled from a failed run that generalizes a wrong diagnosis. A sound item used on the wrong task is B or D, not A. Example: "interpret bulb behavior with caution" makes the agent hunt for circuit faults after a correct negative reading, until the game clock runs out.
- B, a sound item outside its scope. This includes an item written for another task type and applied here, when no other item in memory covers this task better. An item whose scope covers the task is C, not B. Example: melting-point items that say "wait for a plateau" applied to a threshold task, so the agent keeps waiting after the answer is known.
- C, a sound and applicable item that was misused. This includes a procedure adopted as the whole plan that drops the task's own first step. A wrong item is A, not C. Example: a circuit workflow copied step by step that skips the `focus on` step the task requires.
- D, a better item was passed over. The agent followed a sound item while memory held another item that fits this task better. Example: for a threshold task the agent follows the melting-point item "wait for a plateau", although another item says to stop as soon as the thermometer passes the threshold.
- E, no causal link. Give a subtype. `noise` is a slip or a budget pattern the run without memory could show as well. `environment` is a tool or simulator error. `unrelated` is an agent mistake that no item explains.
- unclear. Use it only when a transcript is cut off or the divergence cannot be located.

For every label, also report `better_item`: the item in memory that fits this task better than the one the agent followed and would have avoided the divergence, or null when there is none. It is required for D and may also be set for A.

Confidence is `high`, `medium` or `low`. When torn between two labels, choose one and lower the confidence.

Mark a case exemplary when it would serve as a worked example in a paper: the divergence sits in one turn, a line of reasoning names or paraphrases the item, and the failure follows from it within a few turns. Give the reason in one sentence.
\end{judgebox}
\caption{\textbf{Judge prompt for the causes of degradation.} The case's files follow in the user message. Figure~\ref{fig:regression-criteria-2} continues the prompt.}
\label{fig:regression-criteria}
\end{figure*}

\begin{figure*}[p]
\centering
\begin{judgebox}[System prompt of the regression judge, continued]
# Benchmarks

The transcript header and the `eval_*.json` files state how each episode ended.

- ALFWorld. Success is reaching the goal. The environment allows 49 actions (`budget_exhausted`). `agent_stopped` means the agent ended the episode itself, by sending `done` or ending the session, and the episode counts as failed.
- ScienceWorld. Success is completing the task. The simulator's clock allows 100 game ticks (`game_time_exhausted`). One action can end the task at once with a negative score (`task_failed`), for example focusing on the wrong object. `eval_*.json` lists the subgoals with the step at which each was achieved.
- AppWorld. Success is passing every requirement test. A prose answer on a task whose gold answer is null does not count as a failure. The environment allows 50 code executions. `eval_*.json` lists the passed and failed checks.
- PlanCraft. Success is the target item in the output slot. The environment allows 30 actions. `eval_*.json` gives the target, whether it was reached, and the final inventory.
- tau3-bench retail. Success is a full reward. The agent talks with a simulated user and calls tools over the order database.

# Output

Return one JSON object with exactly these keys.

{"case_id": str, "label": "A"|"B"|"C"|"D"|"E"|"unclear", "confidence": "high"|"medium"|"low", "divergence": {"memory_turn": int or null, "nomem_turn": int or null, "summary": str}, "items": [{"id": str, "quote": str}], "better_item": {"id": str, "quote": str} or null, "evidence": [{"file": str, "kind": str, "quote": str}], "e_subtype": "noise"|"environment"|"unrelated"|null, "failure": str, "rationale": str, "exemplary": bool, "exemplary_reason": str or null}

- `case_id` is the case id given at the top of the user message.
- `items` lists the items involved, each quoted from the file that holds it. Leave it empty for E.
- `better_item` is quoted from `memory.txt`; null when there is none.
- `evidence` quotes the lines that support the label, each from the file it names (`memory_run.md`, `nomem.md`, `memory.txt`, `eval_memory.json` or `eval_nomem.json`), with `kind` one of reasoning, action, observation, memory, evaluation.
- Copy every quote verbatim from its file, at least 12 characters long. A checker compares each quote with the file after collapsing whitespace and refuses the verdict if one is not found.
- `e_subtype` is set for E and null otherwise.
- `failure` names the failed check, the budget or clock that ran out, or the wrong final answer.
- `rationale` is a short paragraph that cites turn numbers.
\end{judgebox}
\caption{\textbf{Judge prompt, continued} from Figure~\ref{fig:regression-criteria}.}
\label{fig:regression-criteria-2}
\end{figure*}

\newcommand{\rlhead}[3]{\noindent{\small\bfseries #1}\hfill{\sffamily\tiny\color{casegrey}#2}\par
  \noindent{\sffamily\tiny\color{casegrey}\textsc{Task}\enspace #3}\par\vspace{3pt}}
\newtcolorbox{rlwhy}[1]{enhanced, colback=caseink!5, colframe=caseink!70, boxrule=0.5pt, arc=1pt, boxsep=0pt,
  left=4pt, right=4pt, top=2.5pt, bottom=2.5pt, before skip=1pt, after skip=5pt,
  fontupper=\normalfont\fontsize{7.6}{9.4}\selectfont\color{caseink}\casepar, before upper={{\sffamily\bfseries Why #1}\enspace}}
\newcommand{\rlnote}[2]{\par\vspace{3pt}{\color{#1!40}\hrule height 0.4pt}\vspace{2.5pt}%
  \noindent{\sffamily\fontsize{6.6}{8}\selectfont\color{#1!85!black}\casepar #2}\par}
\newcommand{\rlsep}{\par\vspace{6pt}{\color{casegrid}\hrule height 0.5pt}\vspace{5pt}}

\begin{figure*}[tp]
\centering
\scriptsize\casepar\setlength{\casetagw}{3.1em}
{\color{caseink}\hrule height 0.8pt}\vspace{4pt}
\rlhead{A\enspace Wrong item, ALFWorld}{ExpeL, trial 2\quad No Memory\,\outok\quad ExpeL\,\outno}{look at vase under the desklamp.}
\begin{rlwhy}{A}The item says the task ends once the lamp is on and the vase is anywhere at the lamp's location. In ALFWorld the agent has to hold the vase, as the source run already shows, so the test run turns on the lamp beside a vase on the desk and declares the task done.\end{rlwhy}
\begin{insightcard}\cardhead{Rule 2}{Lamp tasks finish when the object is next to the lit lamp\enspace\notetag{Wrong item}}{\cardbody Bring the object to the lamp or find it already there, then turn the lamp on. The task is then complete whether the object lies on the surface or is in hand.}\par\vspace{2pt}
{\cardbody\cardfield{Item text}``\ldots{} bring the object to the lamp's location (or ensure it is already there) and turn on the lamp\ldots{} \keyph{the task completes when the object is at the lamp's location (either on the surface or in your hand) and the lamp is on.}''}\end{insightcard}
\begin{rawlane}{Source run (step-0043\,$\cdot$\,success)\hfill\markok}{srcA1}
\rltask{look at pencil under the desklamp.}
\rlturn{T3}{\rla{go to desk 1}\rlo{On the desk 1, you see a book 1, \ldots{} a desklamp 1, \ldots{} a pencil 2, a pencil 1, and a tissuebox 1.}}
\rlturn{T4}{\rlr{\ldots{} Let me first turn on the desklamp and then examine/look at the pencil under it.}\rla{use desklamp 1}\rlo{You turn on the desklamp 1.}}
\rlturn{T5}{\rla{take pencil 1}\rlo{Nothing happens.}}
\begin{rldecisive}\rlturn{T8}{\rlr{Let me try to take the pencil again with proper syntax.}\rla{take pencil 1 from desk 1}\rlo{You pick up the pencil 1 from the desk 1. Episode over\enspace\outok}}\end{rldecisive}
\rlnote{caseink}{The lamp is on from T4 with the pencil on the desk beside it, and the episode goes on through T7. It ends at T8, when the agent picks the pencil up. ExpeL wrote Rule 2 from several successful runs, and this run is in the batch whose edit added ``either on the surface or in your hand''.}
\end{rawlane}\par\vspace{4pt}
\noindent
\begin{minipage}[t]{0.485\textwidth}\vspace{0pt}\begin{withlane}{\markno}{rowA1}
\rlturn{T2}{\rlr{Based on the experience notes, this means I need to bring a vase to the location of a desklamp, and turn on the desklamp.}\rla{look}\rlo{You are in the middle of a room. \ldots{} you see nothing.}}
\rlturn{T3}{\rla{go to desk 1}\rlo{On the desk 1, you see a desklamp 1, a laptop 1, a mug 3, a pencil 2, and a vase 4.}}
\begin{rldecisive}\rlturn{T4}{\rlr{So I have both the vase and the desklamp at the same location. I just need to turn on the desklamp.}\rla{use desklamp 1}\rlo{You turn on the desklamp 1.}}\end{rldecisive}
\rlturn{T5}{\rlr{According to the experience notes, `the task completes when the object is at the lamp's location (either on the surface or in your hand) and the lamp is on.'}\rla{done}\rlo{Episode over\enspace\outno}}
\rlnote{missink}{The vase stays on the desk and is never picked up. Turning on the lamp does not end the episode, and the agent stops on its own.}
\end{withlane}\end{minipage}\hfill
\begin{minipage}[t]{0.485\textwidth}\vspace{0pt}\begin{withoutlane}{\markok}{rowA1}
\rlturn{T4}{\rla{go to shelf 3}\rlo{On the shelf 3, you see a vase 1.}}
\rlturn{T5}{\rlr{Found a vase on shelf 3. Let me take it and then find a desklamp.}\rla{take vase 1 from shelf 3}\rlo{You pick up the vase 1 from the shelf 3.}}
\rlturn{T6}{\rla{go to desk 1}\rlo{On the desk 1, you see a desklamp 1, \ldots{} and a vase 4.}}
\begin{rldecisive}\rlturn{T7}{\rla{use desklamp 1}\rlo{You turn on the desklamp 1. Episode over\enspace\outok}}\end{rldecisive}
\rlnote{helpink}{Holds a vase when it turns on the lamp, and the episode ends in success.}
\end{withoutlane}\end{minipage}\par
\vspace{4pt}{\color{caseink}\hrule height 0.8pt}
\caption{\textbf{Wrong item (A) in ALFWorld.} The block gives the item, its source run and the same test task with and without memory. The source run is the training run the item was written from. A turn (T) shows the reasoning in italics, the action (Act) in typewriter and the observation (Obs) in grey. Shading marks the decisive turn. Item titles, summaries and the coloured notes are ours.}
\label{fig:wrong-items}
\end{figure*}

\begin{figure*}[tp]
\centering
\scriptsize\casepar\setlength{\casetagw}{3.1em}
{\color{caseink}\hrule height 0.8pt}\vspace{4pt}
\rlhead{B\enspace Out-of-scope item, AppWorld}{ReasoningBank, trial 2\quad No Memory\,\outok\quad ReasoningBank\,\outno}{Ronald has sent me a message on phone regarding some amazon order. I am forgetting the details. Do as requested as per their last message.}
\begin{rlwhy}{B}The item is about a contact who asks for information by text and expects a reply, as other AppWorld tasks require. Ronald asks for an Amazon order and a Venmo request and no reply. At T36 the run cites the item and texts him a confirmation, which adds phone records the task did not ask for.\end{rlwhy}
\begin{insightcard}\cardhead{Memory Item 1}{Reply by text with the information a contact asked for\enspace\notetag{Out-of-scope item}}{\cardbody Find the contact, read their request in the text messages, and send a text that answers it.}\par\vspace{2pt}
{\cardbody\cardfield{Description}``\keyph{Use when you need to respond to a text message request from a contact} in the phone app.''\enspace\cardfield{Item text}``First, find the contact by searching the phone contacts. Then search text messages from that contact to read their specific request. Finally, \keyph{send your reply text message with the requested information}, formatted as the contact asked.''}\par\vspace{3pt}
{\cardbody\cardfield{Contrast}Item covers a contact who asks for information by text. Task is an order and a payment request, with no reply asked for.}\end{insightcard}
\noindent
\begin{minipage}[t]{0.485\textwidth}\vspace{0pt}\begin{withlane}{\markno}{rowB}
\rlturn{T11}{\rla{phone.search\_text\_messages(\allowbreak phone\_number="3546176171", \ldots)}\rlo{Ronald, ``Hey, mind ordering these from amazon at your home? \ldots{} Also, please request me the total amount for the order on Venmo. Thanks a lot!''}}
\rlturn{T29}{\rla{amazon.place\_order(\allowbreak address\_id=17, payment\_card\_id=37)}\rlo{Successfully placed the order. \ldots{} 'order\_id': 3146}}
\rlturn{T35}{\rla{venmo.create\_payment\_request(\allowbreak user\_email="ron.harrison@gmail.com", amount=112.25, \ldots)}\rlo{\{"message": "Payment request created."\}}}
\begin{rldecisive}\rlturn{T36}{\noindent\casereason{``Let me also reply to Ronald's text message to let him know what I've done, \keyph{as per the memory item about responding to contact requests.} \ldots{} Let me send a confirmation text as well, since the memory item suggests responding.''}\par\rla{api\_docs.show\_api\_doc(app\_name='phone', \allowbreak api\_name='send\_text\_message')}}\end{rldecisive}
\rlturn{T37}{\rla{phone.send\_text\_message(\allowbreak phone\_number="3546176171", message="Hey Ronald, I ordered the socks from Amazon (order \#3146) \ldots")}\rlo{\{"message": "Text message sent."\}}}
\rlturn{T38}{\rla{supervisor.complete\_task(answer=\ldots)}\rlo{Episode over\enspace\outno}}
\rlnote{missink}{The text adds phone.UserTextMessage and phone.GlobalTextMessage records. The check on model changes allows no phone records and fails.}
\end{withlane}\end{minipage}\hfill
\begin{minipage}[t]{0.485\textwidth}\vspace{0pt}\begin{withoutlane}{\markok}{rowB}
\rlturn{T11}{\rla{phone.search\_text\_messages(\ldots)}\rlo{Ronald, ``Hey, mind ordering these from amazon at your home? \ldots{} Also, please request me the total amount for the order on Venmo. Thanks a lot!''}}
\rlturn{T41}{\rla{amazon.place\_order(\allowbreak payment\_card\_id=37, address\_id=17)}\rlo{Successfully placed the order. \ldots{} "order\_id": 3146}}
\rlturn{T48}{\rla{venmo.create\_payment\_request(\allowbreak user\_email='ron.harrison@gmail.com', amount=112.25, \ldots)}\rlo{\{"message": "Payment request created."\}}}
\begin{rldecisive}\rlturn{T49}{\rlr{Both tasks are done \ldots{} Now I should complete the task.}\rla{supervisor.complete\_task(answer=\ldots)}\rlo{Episode over\enspace\outok}}\end{rldecisive}
\rlnote{helpink}{Does the two things Ronald asked for and sends no text, so the phone app is unchanged.}
\end{withoutlane}\end{minipage}\par
\vspace{4pt}{\color{caseink}\hrule height 0.8pt}
\caption{\textbf{Out-of-scope item (B) in AppWorld.} Laid out as Figure~\ref{fig:wrong-items} without a source run. The item card quotes the item's description and text.}
\label{fig:regression-labels}
\end{figure*}

\begin{figure*}[tp]
\centering
\scriptsize\casepar\setlength{\casetagw}{3.1em}
{\color{caseink}\hrule height 0.8pt}\vspace{4pt}
\rlhead{C\enspace Misused item, $\tau^3$-bench retail}{ReasoningBank, trial 2\quad No Memory\,\outok\quad ReasoningBank\,\outno}{update the address on file and on any orders. After both changes the user asks to revert the address on file.}
\begin{rlwhy}{C}The item is correct. Items on a pending order can be modified only once. At T26 the run stretches it to conclude that other changes, such as the address, can be made several times, which the item does not say. The second address change breaks the database check.\end{rlwhy}
\begin{insightcard}\cardhead{Memory Item 3}{Item changes on a pending order are allowed once, so plan every change first\enspace\notetag{Misused item}}{\cardbody Collect and confirm all item and address changes. Change the address first, then make the one item change, and warn the user that the order cannot change again.}\par\vspace{2pt}
{\cardbody\cardfield{Item text}``For pending orders, \keyph{modifying items can only be done once per order.} Before proceeding, gather all intended modifications (item changes, address updates) and confirm them with the user. First make the address update, then perform the single-allowed item modification, and explicitly warn the user that no further modifications to the order will be possible afterward.''}\end{insightcard}
\noindent
\begin{minipage}[t]{0.485\textwidth}\vspace{0pt}\begin{withlane}{\markno}{rowC}
\rlturn{T19}{\rlr{According to Memory Item 3 \ldots{} But here I'm just modifying the address on the order, not modifying items, so that's fine.}\rla{modify\_user\_address}\rlo{address on file is 101 Highway, New York}}
\rlturn{T20}{\rla{modify\_pending\_order\_address(\allowbreak order\_id='\#W9911714', \ldots)}\rlo{order \#W9911714 ships to 101 Highway, New York}}
\rlturn{User}{\rlr{Actually, I think I made a mistake. Can you change my user address back to the original one in Denver, 80280?}}
\rlturn{T23}{\rla{modify\_user\_address}\rlo{address on file is 667 Highland Drive, Denver}}
\rlturn{Reply}{\rlr{\ldots{} the pending order \#W9911714 still has the 101 Highway, New York address. Would you like me to change that back to the Denver address as well?}}
\rlturn{User}{\rlr{Yes, please switch the shipping address for that order back to the Denver address too.}}
\begin{rldecisive}\rlturn{T26}{\noindent\casereason{``The pending order's address can be modified again since we've only done one modification to it so far and \keyph{there's no restriction on modifying the address multiple times.} The restriction about `once per order' is only for modifying items.''}\par\rla{modify\_pending\_order\_address(\allowbreak order\_id='\#W9911714', \ldots)}\rlo{order \#W9911714 ships to 667 Highland Drive, Denver}}\end{rldecisive}
\rlturn{User}{\rlr{No, that's all for now. Thanks for your help.}\rlo{Episode over\enspace\outno}}
\rlnote{missink}{The run itself offers to revert the order, and the order is modified a second time. The database check expects it to stay at 101 Highway.}
\end{withlane}\end{minipage}\hfill
\begin{minipage}[t]{0.485\textwidth}\vspace{0pt}\begin{withoutlane}{\markok}{rowC}
\rlturn{T20}{\rla{modify\_user\_address}\rlo{address on file is 101 Highway, New York}}
\rlturn{T21}{\rla{modify\_pending\_order\_address(\allowbreak order\_id='\#W9911714', \ldots)}\rlo{order \#W9911714 ships to 101 Highway, New York}}
\rlturn{User}{\rlr{Actually, I think I made a mistake. Can you change my user address back to the original one in Denver?}}
\rlturn{Reply}{\rlr{\ldots{} I can only update the default profile address. The shipping address for order \#W9911714 was already changed to the new address, and I can't undo that since the modify order address tool can only be called once per order.}}
\rlturn{User}{\rlr{Yes, please switch my default address back to the Denver one. That's all I need.}}
\begin{rldecisive}\rlturn{T26}{\rla{modify\_user\_address}\rlo{address on file is 667 Highland Drive, Denver\enspace\outok}}\end{rldecisive}
\rlnote{helpink}{Reverts only the address on file. The order keeps 101 Highway, as the database check expects.}
\end{withoutlane}\end{minipage}\par
\vspace{4pt}{\color{caseink}\hrule height 0.8pt}
\caption{\textbf{Misused item (C) in $\tau^3$-bench retail.} Laid out as Figure~\ref{fig:wrong-items} without a source run. User and Reply rows are messages between the simulated user and the agent. The highlight in turn T26 marks the sentence that stretches the highlighted item phrase.}
\label{fig:regression-labels-c}
\end{figure*}

\begin{figure*}[tp]
\centering
\scriptsize\casepar\setlength{\casetagw}{3.1em}
{\color{caseink}\hrule height 0.8pt}\vspace{4pt}
\rlhead{D\enspace Missed better item, AppWorld}{ExpeL, trial 1\quad No Memory\,\outok\quad ExpeL\,\outno}{add all my coworkers and friends as friends on Venmo, if they are not already.}
\begin{rlwhy}{D}Rule 2 says to page through every list with a page size of 20. The run pages through its Venmo friends but calls \texttt{search\_contacts} once at the default page size of 5. The friend search returns 5 of 6 friends, and William Martinez is never added.\end{rlwhy}
\begin{insightcard}\cardhead{Rule 3}{Find people through the phone app's contact search\enspace\notetag{Followed}}{\cardbody Look a person up with \texttt{search\_contacts} and a query or a relationship filter.}\par\vspace{2pt}
{\cardbody\cardfield{Item text}``To locate information about a specific person (contact, email, phone number), \keyph{use the phone app's \texttt{search\_contacts} with a query or relationship filter.}''}\par\vspace{4pt}
\cardhead{Rule 2}{Page through every list with page\_limit 20\enspace\oktag{Better item}}{\cardbody Call a paginated API in a loop over \texttt{page\_index} until a page comes back empty or short.}\par\vspace{2pt}
{\cardbody\cardfield{Item text}``When using any app that returns paginated lists \ldots, \keyph{implement a while loop that increments \texttt{page\_index}}, stops when the response is empty or when the returned list length is less than the \texttt{page\_limit}, and \keyph{use a \texttt{page\_limit} of 20} to minimize calls.''}\par\vspace{3pt}
{\cardbody\cardfield{API doc}\texttt{search\_contacts} takes \texttt{page\_limit} with a default of 5. The run with memory never reads this doc, because its request at T7 is rejected and not repeated.}\end{insightcard}
\noindent
\begin{minipage}[t]{0.485\textwidth}\vspace{0pt}\begin{withlane}{\markno}{rowD}
\rlturn{T7}{\rla{api\_docs.show\_api\_doc(app\_name='phone', \allowbreak api\_name='search\_contacts')}\rlo{one exec per turn, read this output before sending the next chunk}}
\rlturn{T14}{\rla{phone.show\_contact\_relationships()}\rlo{["coworker", "father", "friend", "manager", \ldots]}}
\begin{rldecisive}\rlturn{T15}{\rlr{So I need to find contacts with relationship "coworker" and "friend". Let me search for contacts with these relationships.}\rla{phone.search\_contacts(\allowbreak relationship='coworker')}\rla{phone.search\_contacts(\allowbreak relationship='friend')}\rlo{\keyph{5 coworkers and 5 friends}, one page each}}\end{rldecisive}
\rlturn{T16}{\rlr{Friends: Cheryl Martinez, Cody Smith, Cory Wilson, Grant Martinez, Travis Wilson}}
\rlturn{T21}{\rlr{Get current friends - paginate through all}\rla{while True: venmo.search\_friends(\allowbreak page\_index=page\_idx, \allowbreak page\_limit=20) \ldots{} page\_idx += 1}\rlo{the user's current Venmo friends, all pages}}
\rlturn{T23}{\rla{venmo.add\_friend for Andrew, Chelsea, Eric, Nicholas, Cheryl and Cody}\rlo{Added to your friend list. (six times)}}
\rlturn{T24}{\rla{supervisor.complete\_task(answer="Added 6 friends \ldots")}\rlo{Episode over\enspace\outno}}
\rlnote{missink}{Pages through the Venmo friends but not through the contacts. The three coworkers it misses are already Venmo friends, so the one missing add is William Martinez.}
\end{withlane}\end{minipage}\hfill
\begin{minipage}[t]{0.485\textwidth}\vspace{0pt}\begin{withoutlane}{\markok}{rowD}
\rlturn{T21}{\rla{phone.show\_contact\_relationships()}\rlo{["coworker", "father", "friend", "manager", \ldots]}}
\rlturn{T22}{\rla{while True: phone.search\_contacts(\allowbreak relationship="coworker", \allowbreak page\_index=page\_index, \allowbreak page\_limit=20) \ldots}\rlo{Coworkers: \keyph{8}}}
\begin{rldecisive}\rlturn{T23}{\rla{while True: phone.search\_contacts(\allowbreak relationship="friend", \allowbreak page\_index=page\_index, \allowbreak page\_limit=20) \ldots}\rlo{Friends: \keyph{6}. Cheryl Martinez, Cody Smith, Cory Wilson, Grant Martinez, Travis Wilson, William Martinez}}\end{rldecisive}
\rlturn{T32}{\rla{while True: venmo.search\_friends(\allowbreak page\_index=page\_index, \allowbreak page\_limit=20) \ldots}\rlo{Current Venmo friends: 9}}
\rlturn{T34}{\rla{venmo.add\_friend for seven people, including william\_mart@gmail.com}\rlo{Added to your friend list. (seven times)}}
\rlturn{T36}{\rla{supervisor.complete\_task(answer=\ldots)}\rlo{Episode over\enspace\outok}}
\rlnote{helpink}{Pages through the contact search as Rule 2 says and finds all six friends.}
\end{withoutlane}\end{minipage}\par
\vspace{4pt}{\color{caseink}\hrule height 0.8pt}
\caption{\textbf{Missed better item (D) in AppWorld.} Laid out as Figure~\ref{fig:wrong-items} without a source run. Rules are numbered as in the memory.}
\label{fig:regression-labels-d}
\end{figure*}

\begin{figure*}[tp]
\centering
\scriptsize\casepar\setlength{\casetagw}{3.1em}
{\color{caseink}\hrule height 0.8pt}\vspace{4pt}
\rlhead{E\enspace No link to memory, AppWorld}{ExpeL, trial 0\quad No Memory\,\outok\quad ExpeL\,\outno}{send \$250 on Venmo to Catherine.}
\begin{rlwhy}{E}The run adds \$250 to the Venmo balance from a newly added card before paying. That top-up is a BankTransfer the task did not ask for, and it fails the check on model changes. None of the 12 rules suggests a top-up.\end{rlwhy}
\begin{insightcard}\cardhead{No item}{None of the 12 rules bears on the top-up\enspace\notetag{No link to memory}}{\cardbody The rules cover API exploration, dates, credentials, pagination, aggregation, replies and file export. None of them says to add money to the Venmo balance before paying.}\end{insightcard}
\noindent
\begin{minipage}[t]{0.485\textwidth}\vspace{0pt}\begin{withlane}{\markno}{rowE}
\rlturn{T17}{\rla{venmo.create\_transaction(\allowbreak receiver\_email="ca-smit@gmail.com", amount=250.0, \ldots)}\rlo{Your Venmo balance does not have \$250.00 to make this transaction.}}
\rlturn{T21}{\rla{venmo.add\_to\_venmo\_balance(\allowbreak amount=250.0, payment\_card\_id=206)}\rlo{The payment card has expired.}}
\rlturn{T22}{\rlr{The Discover card has expired (expiry\_year 2022, expiry\_month 11 - current date is 2026-10-07). \ldots{} All cards seem to be expired based on the current date 2026-10-07.}}
\rlturn{T26}{\rla{venmo.add\_payment\_card(card\_name="Wells Fargo", \ldots, expiry\_year=2023, expiry\_month=9, \ldots)}\rlo{Added payment card to your account. payment\_card\_id 478}}
\begin{rldecisive}\rlturn{T27}{\rlr{The card was added. But it's also expired (expiry 2023/9). Let me try adding money to my balance using it and see what happens.}\rla{venmo.add\_to\_venmo\_balance(\allowbreak amount=250.0, payment\_card\_id=478)}\rlo{\{"message": "Money added to Venmo balance.", "bank\_transfer\_id": 1\}}}\end{rldecisive}
\rlturn{T29}{\rla{venmo.create\_transaction(\allowbreak receiver\_email="ca-smit@gmail.com", amount=250.0, \ldots)}\rlo{Sent money.}}
\rlturn{T30}{\rla{supervisor.complete\_task(answer="success")}\rlo{Episode over\enspace\outno}}
\rlnote{missink}{The top-up adds a venmo.BankTransfer record. The check expects changes only to Transaction, Notification and User records and fails.}
\end{withlane}\end{minipage}\hfill
\begin{minipage}[t]{0.485\textwidth}\vspace{0pt}\begin{withoutlane}{\markok}{rowE}
\rlturn{T30}{\rla{venmo.create\_transaction(\ldots, payment\_card\_id=card\_id) for cards 206, 208, 209, 210}\rlo{The payment card has expired.}}
\rlturn{T34}{\rlr{So card 206 (Discover) and 210 (Chase) are expired (2022 expiry). Cards 208 (MasterCard, expires 2023) and 209 (HSBC, expires 2023) don't have \$250.}}
\rlturn{T35}{\rla{venmo.add\_payment\_card(card\_name="Wells Fargo", \ldots, expiry\_year=2023, expiry\_month=9, \ldots)}\rlo{Added payment card to your account. payment\_card\_id 478}}
\begin{rldecisive}\rlturn{T36}{\rlr{The Wells Fargo card was added with payment\_card\_id=478. Now let me try using this card to send the money.}\rla{venmo.create\_transaction(\allowbreak receiver\_email="ca-smit@gmail.com", amount=250.0, payment\_card\_id=478, \ldots)}\rlo{\{"message": "Sent money.", "transaction\_id": 8216\}}}\end{rldecisive}
\rlturn{T37}{\rla{supervisor.complete\_task(answer=\ldots)}\rlo{Episode over\enspace\outok}}
\rlnote{helpink}{Pays from the new card directly, so no BankTransfer is created.}
\end{withoutlane}\end{minipage}\par
\vspace{4pt}{\color{caseink}\hrule height 0.8pt}
\caption{\textbf{No link to memory (E) in AppWorld.} Laid out as Figure~\ref{fig:wrong-items} without a source run. The world clock of the task reads 2023-05-18, and the date the run quotes in T22 comes from its own prompt.}
\label{fig:regression-labels-e}
\end{figure*}

\clearpage
\section{How WikiTrace Uses Its Evidence}
\label{sec:appendix-evidence}

\subsection{Descent Judge}
\label{sec:appendix-descent-judge}

The judge behind Figures~\ref{fig:descent-before} and~\ref{fig:descent-after} reads one WikiTrace run in a text-world benchmark and labels every descent in it, where a descent is a stretch of consecutive memory calls that opens a compilation or a raw run and ends at the next game action. It receives the system prompt of Figure~\ref{fig:descent-judge} and three user messages: the insights the agent saw, the agent's full transcript with each descent marked, and the prompt of one axis (Figures~\ref{fig:descent-judge-2} and~\ref{fig:descent-judge-3}). Each axis is a separate call. The trigger and the target are read from the agent's reasoning before the descent, and the effect from what the agent does after it. Every label carries a verbatim quote from the agent's messages, and a checker sends an answer back once if a quote is not found there.

\texttt{gpt-5.6-sol} labels the trigger and the target, and \texttt{GLM-5.3} labels them too, agreeing at Cohen's $\kappa$ of 0.77 and 0.63. The two judges agree on the effect only at $\kappa = 0.46$, so the effect label is the majority of these two and \texttt{Claude Opus 5.5} (Fleiss' $\kappa = 0.49$). All three judges run at medium reasoning effort. Of the 508 descents, 46 have no majority. The judges label 508 descents in the 430 descending runs; one descent has no trigger or target because its answer failed the checker twice. The prompt calls a \dlabel{blocked attempt} a \texttt{setback}; the other labels in the text are the prompt's codes.

\begin{figure*}[p]
\centering
\begin{judgebox}[System prompt of the descent judge]
You code one aspect of why and how an LLM agent consulted its memory of past runs while playing a text game. That aspect, the axis, is given in the last message, after the transcript.

## The agent and its memory

The agent plays one text game episode: an ALFWorld household task, a ScienceWorld science task, or a PlanCraft crafting task. Its memory holds three levels built from earlier runs on other tasks:

- insights: short lessons, each naming the run it came from as its source. All insights are in the agent's first message.
- compilations: one summary per past run.
- raw runs: the full log of each past run (its actions, observations and the agent's reasoning). The agent can list a run's actions, read its turns, read files, or grep across runs.

The first message also lists a few past runs as similar to the current task.

## What you receive

1. Insights the agent saw: the insights block of the agent's first message, shared by every case of the same benchmark and trial.
2. The transcript: the agent's own context, unclipped. It starts with the agent's system prompt, then every message in order, numbered `-- message N - <role> --`. In an agent message, `thinking |` is its private reasoning, `text |` is text it wrote, and `call #n |` is a tool call. Each call's result follows as its own message. Game actions are `act` or `take_action` calls; every other call reads the memory. In message 0 the insights are replaced by a pointer to item 1.
3. Burst marks. A burst is one stretch of consecutive memory calls that opens past runs: it starts at the first call that opens a compilation or a raw run and ends at the agent's next game action or message. `======== BURST k START ========` and `======== BURST k END ========` mark each burst. Under each START line, a bracketed line computed from the calls lists the runs the burst opens, with their outcomes, and the insights whose source is one of those runs.
4. The axis: its codes, how to decide among them, and the evidence to give.

Read the whole transcript. Code every marked burst, in order, on this axis alone.

## Evidence

For each burst give one verbatim quote and the number of the agent message it comes from. Copy the quote exactly, at least 12 characters, from a single message. Return only the JSON object the schema asks for, with one entry per marked burst, in order.
\end{judgebox}
\caption{\textbf{Descent judge system prompt.} The insights, the transcript and one axis prompt (Figures~\ref{fig:descent-judge-2} and~\ref{fig:descent-judge-3}) follow as three user messages.}
\label{fig:descent-judge}
\end{figure*}

\begin{figure*}[p]
\centering
\begin{judgebox}[Axis prompts: trigger and target]
## Axis: trigger, why the agent opened past runs at this point

Code the reason the agent states for opening the runs, in the message that makes the burst's first opening call and the messages before it.

| code | definition | against its nearest neighbour | |---|---|---| | setback | An observation in this episode blocked or contradicted the agent's plan, and the agent treats it as a problem: an action failed or had no effect, a result differed from what the agent expected, a search in the game found nothing, or a needed item turned out to be missing. | Against open question: there is a concrete observation that the agent names as the problem. Uncertainty about the next step with nothing gone wrong is open question, and so is a result the agent does not treat as a problem. | | insight doubt | The agent questions whether an insight holds, because two insights disagree or because the insight came from a failed run, and opens runs to settle it. | Against failed precedent: the agent names an insight, or says the insights disagree, and questions whether it holds. Mentioning an insight without questioning it is not insight doubt. Opening a failed run only because it failed, with no doubted insight, is failed precedent. | | failed precedent | A past run listed as similar to the current task failed, and the agent opens it to avoid repeating that failure; nothing has gone wrong in its own episode. | Against routine check: the opened run is a failed run and the agent says it wants to avoid what went wrong there. | | open question | The agent names one specific thing it does not know about its next step, and nothing has gone wrong. | Against routine check: the agent states a specific question or names alternatives. A general wish to see how a run did the task is routine check. | | routine check | The agent opens a run that matches its task, usually a successful one, to see how that run did it, as a precaution and without a stated problem or question. | The default when none of the above applies. |

Use the order setback > insight doubt > failed precedent > open question > routine check only when the agent gives two reasons.

## Fields

- code: the trigger code.
- quote and msg: the agent's own words (thinking or text) that show why it opened the runs, and their message number.
- note: one sentence on how you decided.

## Axis: target, what the agent was looking for in the runs it opened

Code what the agent says it wants from the runs, in the message that makes the burst's first opening call and the messages before it. If its thinking and its text differ, follow its thinking.

| code | definition | against its nearest neighbour | |---|---|---| | workaround | A way past something that blocks the agent: another way to make or obtain what it needs, where a missing item can be found, or how another run got past the same block. Often searched for by content across runs, because no listed run is known to cover it. | Against specific fact: a workaround search is open-ended, looking for any run that might show a way through. A search that names one question to answer is specific fact, even when it looks across many runs. | | failure cause | What went wrong in a failed past run. | Against workaround: the agent wants the failure explained, not a way out of its own block. If its own episode is already blocked and it asks how the failed run got past the same block, use workaround. Against specific fact: opening the failed run an insight came from to ask whether the insight holds is specific fact; asking what went wrong in that run is failure cause. | | specific fact | The answer to one named question: how a command is written, whether something works or is needed, whether an insight holds, how a result appears, or how the game behaves. | Against procedure: it is driven by one question. Reading a whole sequence with one part singled out stays procedure. | | procedure | The working sequence of a run: the order of its actions, often with where things were found. | The default for reading a run to reproduce its steps. |

If the agent names two targets: workaround > failure cause > specific fact > procedure.

## Fields

- code: the target code.
- quote and msg: the agent's own words, or a memory call, that show what it looked for, and their message number.
- note: one sentence on how you decided.
\end{judgebox}
\caption{\textbf{Descent judge axis prompts for trigger and target}, each sent in its own call.}
\label{fig:descent-judge-2}
\end{figure*}

\begin{figure*}[p]
\centering
\begin{judgebox}[Axis prompt: effect]
## Axis: effect, how the opened runs affected the agent's next actions

Find each point where the agent draws on the opened runs after this burst's END and before the next START, or the transcript's end: it says it took information from them, or acts on information only they supply. Code its adopted interpretation, including mistaken interpretations, not thoughts it rejects. Use need not succeed.

The baseline is all insights, the current system prompt and its examples, the task, observations available before this burst even if first noticed through a run, the stated plan, and information adopted from earlier bursts. Earlier mention alone does not establish adoption. The system prompt printed at the start of an opened run's log is baseline only insofar as its information was already available.

Compare only against the insights the agent names, in thinking or text, from the start of the episode to the deciding point, and the insights listed under this burst's START line as sourced from the opened runs; do not search the other insights. Compare the specific information used, not merely the action's general purpose. Do not attribute discoveries from current game results to the run. Respect each insight's conditions and scope.

| code | definition | against its nearest neighbour | |---|---|---| | revise insight | Because of the run, the agent acts on a correction to an applicable insight's rule, condition or scope, or acts otherwise than that insight directs. | Against confirm plan and refine insight: check every insight in that set, not only the one followed. Revision takes priority even if another insight states the chosen action and the displaced insight does not rule it out. Choosing between conflicting prescriptions is revision. | | refine insight | The agent uses a run-supplied detail about a step an insight addresses, or uses the run to settle an unresolved choice among its listed options. | Against revise insight: no applicable insight directs otherwise. Against confirm plan: refine insight only if the insight leaves something open that the agent must decide (which object, where, in what order, what value, how a command is written) and the agent decides it from the run; if the insight already fixes it, confirm plan. A fact true only of the opened run's episode is not a detail unless the agent shows it also holds in this episode. A choice already settled by the baseline, or an option demonstrated in the current system examples, is baseline. | | new fact | The agent uses run-supplied information absent from the baseline about a step no insight addresses. | Against refine insight: identify the step, not merely the broad task. If an insight addresses that step and the run supplies its missing detail, use refine insight. | | confirm plan | The agent explicitly says the run agrees with its plan or an insight, and the information taken is already in the baseline. | Against none: matching actions alone do not establish stated agreement. Against revise insight: a conflicting applicable prescription takes priority. | | none | No qualifying run-derived use or explicit confirmation occurs. | Includes rejected or unused information, and baseline-only use without stated agreement. |

Code every qualifying point and give the burst the highest: revise insight > refine insight > new fact > confirm plan > none.

## Fields

- code: the effect code.
- quote and msg: the agent's verbatim words or game action at the deciding point, and its message number. For none, quote its rejection or next game action.
- note: one sentence identifying the information used and its baseline relationship; name the relevant insight when revising or refining.
\end{judgebox}
\caption{\textbf{Descent judge axis prompt for effect.}}
\label{fig:descent-judge-3}
\end{figure*}

\subsection{How a Descent Helps}
\label{sec:appendix-descent-cases}

We read pairs of episodes, WikiTrace and Insights Only on the same task and trial, whose outcomes differ, and give a \dlabel{revise insight} in ScienceWorld and a \dlabel{new fact} in ALFWorld (Figure~\ref{fig:descent-cases}).

\newcommand{\dxhead}[3]{\noindent{\normalsize\bfseries #1}\par\vspace{1pt}
  \noindent{\sffamily\tiny\color{casegrey}#2}\par
  \noindent{\sffamily\tiny\color{casegrey}\textsc{Task}\enspace #3}\par\vspace{3pt}}
\begin{figure*}[t]
\centering
\scriptsize\casepar\setlength{\casetagw}{0pt}
{\color{caseink}\hrule height 0.8pt}\vspace{4pt}
\noindent\begin{minipage}[t]{0.488\textwidth}\vspace{0pt}
\dxhead{Revise Insight (success)}{ScienceWorld, trial 0\quad Insights Only\,\outno\quad WikiTrace\,\outok}{determine if unknown substance V is electrically conductive, blue box if it is, orange box if not.}
\begin{dxwhy}Insights drawn from failed runs say to always put the switch in the circuit. Built that way, the circuit stays dark even without the substance, so the bulb cannot tell whether V conducts.\end{dxwhy}
\begin{insightcard}\cardhead{insight-0080}{Use the switch to complete the circuit\enspace\notetag{Wrong}}{\cardbody Always include the switch in series and turn it on.}\end{insightcard}
\begin{rawlane}{Opened run (step-0007\,$\cdot$\,success)\hfill\markok}{dxsrc}
\rltask{determine if aluminum foil is electrically conductive.}
\rlturn{}{\rla{connect battery anode to black wire terminal 1}}
\rlturn{}{\rla{connect black wire terminal 2 to cathode in red light bulb}}
\rlturn{}{\rla{connect aluminum foil terminal 2 to anode in red light bulb}}
\rlturn{}{\rla{move aluminum foil to blue box}}
\dxnote{caseink}{The successful run wires the bulb and the object in one loop with \keyph{no switch}. The task's own example circuit has no switch either, and both agents see it.}
\end{rawlane}\par\vspace{3pt}
\begin{wtlane}{\markok}{dxwt}
\rlturn{}{\rlr{Looking at the past failed runs, they mention: 1. Include the switch in the circuit (insight-0080, insight-0140) \ldots}}
\rlturn{}{\rlr{Let me also look at the raw wiki files for one of the successful conductivity test runs to understand the correct wiring order.}}
\rlturn{}{\rla{/\allowbreak{}work/\allowbreak{}tools/\allowbreak{}turn step-0007 10 20}\enspace\dxdescendrun}
\begin{rldecisive}\rlturn{}{\rlr{The bulb didn't light even when I bypassed unknown V. So something was wrong with my circuit. Let me try without the switch, following the exact pattern from step-0007.}}\end{rldecisive}
\rlturn{}{\rla{connect unknown substance V terminal 2 to anode in red light bulb}\rlo{a red light bulb, which is on.}}
\rlturn{}{\rla{move unknown substance V to blue box}}
\dxnote{caseink}{Drops the switch the insights demand and answers correctly.}
\end{wtlane}\par\vspace{3pt}
\begin{kilane}{\markno}{dxki}
\rlturn{}{\rlr{From the similar past runs, I see: \ldots{} 2. Include the switch in the circuit and turn it on.}}
\rlturn{}{\rla{turn on switch}\rlo{a yellow light bulb, which is off.}}
\rlturn{}{\rlr{The bulb was on when I bypassed the unknown substance V, but it's off when V is in the circuit. This means unknown substance V is nonconductive.}\rla{move unknown substance V to orange box}}
\dxnote{caseink}{Keeps the switch as the insights say, reads the dark bulb as nonconductive and answers wrong.}
\end{kilane}
\end{minipage}\hfill
\begin{minipage}[t]{0.488\textwidth}\vspace{0pt}
\dxhead{New Fact (success)}{ALFWorld unseen, trial 2\quad Insights Only\,\outno\quad WikiTrace\,\outok}{clean some cloth and put it in cabinet.}
\begin{dxwhy}The insight sends the agent to a towel holder, where it finds a handtowel. It does not say that a cloth is an object of its own, kept in a cabinet, so the handtowel passes for the cloth, and cleaning it does nothing.\end{dxwhy}
\begin{insightcard}\cardhead{insight-0117}{Check dedicated holders first\enspace\notetag{Incomplete}}{\cardbody When searching for an object like a handtowel, first check its holder rather than containers like cabinets.}\end{insightcard}
\begin{rawlane}{Source run (step-0042\,$\cdot$\,success)\hfill\markok}{dxsrc}
\rltask{put some handtowel on cabinet.}
\rlturn{}{\rla{open cabinet 7}\rlo{You open the cabinet 7. The cabinet 7 is open. In it, you see a cloth 1.}}
\dxnote{caseink}{In the run, a \keyph{cloth is an object of its own}, found in a cabinet apart from the handtowel.}
\end{rawlane}\par\vspace{3pt}
\begin{wtlane}{\markok}{dxwt}
\rlturn{}{\rla{take handtowel 1 from handtowelholder 1}\rla{clean handtowel 1 with sinkbasin 1}\rlo{Nothing happens.}}
\rlturn{}{\rlr{Maybe ``cloth'' is a specific object type, not a handtowel. Let me search for ``cloth'' in the raw data.}}
\rlturn{}{\rla{grep "cloth" /\allowbreak{}work/\allowbreak{}corpus/\allowbreak{}raw/\allowbreak{}step-0042/\allowbreak{}solver.full.jsonl}\enspace\dxdescend}
\begin{rldecisive}\rlturn{}{\rlr{So ``cloth'' is a specific object type in this environment, not a handtowel or towel. \ldots{} I need to find ``cloth'' -- let me check the cabinets.}}\end{rldecisive}
\rlturn{}{\rla{open cabinet 2}\rlo{In it, you see a cloth 1.}}
\rlturn{}{\rla{clean cloth 1 with sinkbasin 1}\rla{move cloth 1 to cabinet 2}\rlo{Episode over}}
\dxnote{caseink}{Learns from the run that a cloth is its own object kept in cabinets, finds one and cleans it.}
\end{wtlane}\par\vspace{3pt}
\begin{kilane}{\markno}{dxki}
\rlturn{}{\rlr{I found a handtowel 1 on handtowelholder 1. A handtowel is a cloth.}\rla{take handtowel 1 from handtowelholder 1}}
\rlturn{}{\rla{clean handtowel 1 with sinkbasin 1}\rlo{Nothing happens.}}
\rlturn{}{\rlr{\ldots{} maybe the handtowel is already clean and the task just considers it clean.}}
\rlturn{}{\rla{move handtowel 1 to cabinet 1}\rla{done}}
\dxnote{caseink}{Takes the handtowel for the cloth, cannot clean it and puts it away.}
\end{kilane}
\end{minipage}\par
\vspace{4pt}{\color{caseink}\hrule height 0.8pt}
\caption{\textbf{A revised insight and a new fact in full.} Each case gives the task, why its insight cannot be used as it stands, the run WikiTrace opened, and what WikiTrace and Insights Only then do. Reasoning is in italics, actions in typewriter and observations in grey. Shading marks the decisive reasoning. Insight summaries and the coloured notes are ours.}
\label{fig:descent-cases}
\end{figure*}

\clearpage
\section{Prompt Templates}
\label{sec:appendix-prompts}

We give the prompts of the current implementation, those of WikiTrace first. Slate boxes hold the compiler prompts, shared by every benchmark (Figures~\ref{fig:prompt-compiler-success-system} and~\ref{fig:prompt-compiler-failure-system}). Green boxes hold the solver prompts. Figures~\ref{fig:prompt-memory-wikitrace} and~\ref{fig:prompt-memory-controls} give what WikiTrace and Insights Only add to the no-memory solver prompt, which every method starts from, and Figures~\ref{fig:prompt-solver-alfworld} to~\ref{fig:prompt-solver-appworld-2} give that prompt for each benchmark. On $\tau^3$-bench and AppWorld, WikiTrace also changes a few lines of it, marked as in a diff, with removed lines in red and added lines in green. Angle brackets mark slots, including the exemplars we elide.

\begin{figure*}[p]
\centering
\begin{promptbox}[Compiler system prompt, successful episode]
You are an expert in <task kind>. You will be given a task, the corresponding
trajectory with observations that represents **how an agent successfully
accomplished the task**, and the system prompt the agent ran under (its worked
examples are omitted).

Write every field of your output in English only.

<solver_prompt>
<the solver's system prompt, without its exemplars>
</solver_prompt>

## Guidelines
You need to extract and summarize useful insights in the format of candidate
insights based on the agent's successful trajectory, together with the key
decisions the insights rest on.
The goal of candidate insights is to be helpful and generalizable for future
similar tasks.

## Important notes
  - You must first think why the trajectory is successful, and then summarize
    the insights.
  - A key decision is a choice that was necessary for success. Its support is
    the observation that shows the decision was necessary. Not every action
    was a decision.
  - You can extract *at most 3* candidate insights from the trajectory.
  - You must not repeat similar or overlapping insights.
  - Prefer concrete, actionable procedures over abstract principles.
    Name the kinds of objects, the items, apps or APIs a procedure uses. Do not
    copy numbered instances, ids or values that hold only in this one episode.

## Output Format
Your output must strictly follow the JSON format shown below. English only.

{
    "title": <title of the case>,
    "description": <one sentence summary describing the case>,
    "key_decisions": [<key decisions made by the agent>],
    "candidate_insights": [<candidate insights>]
}

Each key decision is a dict following the format shown below.

{
    "decision": <what the agent chose to do, at which step>,
    "support": <what happened next in the trajectory that shows this choice was necessary for success>
}

Each candidate insight is a dict following the format shown below.

{
    "title": <the insight in a few words>,
    "guideline": <what to do in one or two sentences>,
    "applicability": <when this applies, and when it does not>
}
\end{promptbox}
\caption{\textbf{Compiler system prompt for a successful training episode.} Angle brackets mark slots filled per benchmark or by the model.}
\label{fig:prompt-compiler-success-system}
\end{figure*}

\begin{figure*}[p]
\centering
\begin{promptbox}[Compiler system prompt, failed episode]
You are an expert in <task kind>. You will be given a task, the corresponding
trajectory with observations that represents **how an agent attempted the task
but failed**, and the system prompt the agent ran under (its worked examples
are omitted).

Write every field of your output in English only.

<solver_prompt>
<the solver's system prompt, without its exemplars>
</solver_prompt>

## Guidelines
You need to extract and summarize useful insights in the format of candidate
insights based on the agent's failed trajectory, together with the decisions
that caused the failure.
The goal of candidate insights is to be helpful and generalizable for future
similar tasks.

## Important notes
  - The environment did not report the task done. It may have completed most
    of it. The outcome does not say how far it got.
  - The outcome block says how the episode ended. An ending by the harness is
    not a decision the agent made.
  - You must first reflect and think why the trajectory failed, and then
    summarize what lessons you have learned or strategies to prevent the
    failure in the future.
  - A failure cause is a decision the agent made, or did not make. Its support
    is the observation that shows the decision went wrong.
  - You can extract *at most 3* candidate insights from the trajectory.
  - You must not repeat similar or overlapping insights.
  - Prefer concrete, actionable recovery procedures over abstract principles.
    Name the kinds of objects, the items, apps or APIs a procedure uses. Do not
    copy numbered instances, ids or values that hold only in this one episode.

## Output Format
Your output must strictly follow the JSON format shown below. English only.

{
    "title": <title of the case>,
    "description": <one sentence summary describing the case>,
    "failure_causes": [<decisions that caused the failure>],
    "candidate_insights": [<candidate insights>]
}

Each failure cause is a dict following the format shown below.

{
    "decision": <what the agent chose to do, or not to do, at which step>,
    "support": <what happened next in the trajectory that shows this choice caused the failure>
}

Each candidate insight is a dict following the format shown below.

{
    "title": <the insight in a few words>,
    "guideline": <what to do in one or two sentences>,
    "applicability": <when this applies, and when it does not>
}
\end{promptbox}
\begin{promptbox}[Compiler user prompt]
<task>
<the task, as the solver received it>
</task>

<trajectory>
<the episode's steps, each with reasoning, action, and observation>
</trajectory>

<execution_outcome>
<success or failure, the number of steps, and how the episode ended>
</execution_outcome>

Compile this <successful|failed> trajectory.
\end{promptbox}
\caption{\textbf{Compiler system prompt for a failed training episode, and the compiler user prompt shared by both outcomes.}}
\label{fig:prompt-compiler-failure-system}
\end{figure*}

\begin{figure*}[p]
\centering
\begin{solverbox}[WikiTrace system block]
You also have insights from past train runs, inlined at the start of the task message, and
a wiki at /work/corpus, read-only, that holds the runs they came from. They are not the
answer to this task. The task message lists the past runs most similar to this task.

The wiki has three layers.
- /work/corpus/raw/step-NNNN/ is one past run as it happened.
    metadata.md        its task and outcome
    solver.full.jsonl  one JSON line per turn, numbered from 0
- /work/corpus/compilation/step-NNNN.md is a model's account of that run. The model saw
  only whether the run succeeded or failed. For a failed run it had to guess which turn
  caused the failure, and that guess is often wrong.
- The insights in the task message were drawn from the compilations. One from a failed run
  inherits that guess.
Only the raw turns are certain. Each layer has an _index.md.

Reading the wiki with `read` or `bash` costs no game steps. Go down to an insight's source
run and compilation when it tells you what to do but not exactly how, when it came from a
failed run, or when two insights disagree. Check this before your first `act` for the
insights you will rely on, and again whenever an action fails. Look for the one fact that
settles it, such as a command format, in the run. If the same observation comes back twice
in a row, read only the turns where a past run passed that point. Find runs through an
insight's source run, the lower layers' indexes, or a search by the object or the
receptacle. Use the `bash` tool freely:
  grep -rl "mug" /work/corpus/compilation /work/corpus/raw | head -n 20
  for s in 0002 0017; do cat /work/corpus/compilation/step-$s.md; done
  /work/tools/actions step-0017      # every turn's action, one line each
  /work/tools/turn step-0017 7 9     # turns 7 to 9 in full: reasoning, action, observation

Memory rules
- Never replay a past run. It started from a different state than this task, even when
  its task reads the same, and steps that worked there fail here.
- Take from a run only a fact you can check against this task, such as a command format.
  Name what differs before you use it.
- From a failed run, take only the action to avoid, never the plan.
\end{solverbox}
\begin{solverbox}[WikiTrace user block]
<memory_documents>
<insight id="insight-0003">
- **Title:** Activate light source before examination
- **Task type:** look_at_obj_in_light
- **Insight:** When a task involves examining an object with a light (e.g., a desklamp), locate and turn on the light source first, before bringing the object to it. The light must be lit for the examination to succeed.
- **Applicability:** Applies to any task that requires using a light source to examine an object; may not apply if the light source is always on or the light is not necessary.
- **Source:** step-0001 (success)
</insight>
<140 more insights, by task type, then id>
</memory_documents>

<the no-memory user prompt>

<similar_past_runs>
The past train runs most similar to this task, with the insights drawn from each. The
insights themselves are at the start of this message.

### step-0012 (success)
Task: Your task is to: put some book on sidetable.
Insights from this run
- insight-0035: Systematic Search of All Furniture
- insight-0036: Check Beds for Small Objects
- insight-0037: Choose Any Matching Destination When Multiple Exist

<two more past runs in the same form>
</similar_past_runs>
\end{solverbox}
\begin{solverbox}[WikiTrace citation prompt]
The episode is over. Do not call any tool. Output a single JSON object and nothing
else, no markdown and no code fence. Each key holds a list of the materially-used ids,
possibly empty:
- `R`: raw trajectories, e.g. ["step-0023"]
- `C`: compilations, e.g. ["step-0023"]
- `I`: insights, e.g. ["insight-0040"]
Cite only what affected your actions. Do not cite an item merely because it appeared.
\end{solverbox}
\caption{\textbf{WikiTrace additions to the ALFWorld solver prompt.} The system block precedes the runtime notes, and the citation prompt follows the episode.}
\label{fig:prompt-memory-wikitrace}
\end{figure*}

\begin{figure*}[p]
\centering
\begin{solverbox}[Insights Only system block]
You also have insights from past train runs, inlined at the start of the task message.
They are not the answer to this task. The task message lists the past train runs most
similar to this task and the insights drawn from each.
\end{solverbox}
\begin{solverbox}[Insights Only user block]
<memory_documents>
<insight id="insight-0003">
- **Title:** Activate light source before examination
- **Task type:** look_at_obj_in_light
- **Insight:** When a task involves examining an object with a light (e.g., a desklamp), locate and turn on the light source first, before bringing the object to it. The light must be lit for the examination to succeed.
- **Applicability:** Applies to any task that requires using a light source to examine an object; may not apply if the light source is always on or the light is not necessary.
</insight>
<140 more insights, by task type, then id>
</memory_documents>

<the no-memory user prompt>

<the similar_past_runs block of WikiTrace's user block>
\end{solverbox}
\caption{\textbf{Insights Only additions to the ALFWorld solver prompt.} The blocks sit where WikiTrace's do. The insights are listed as in WikiTrace's user block, without their source runs. Nothing is mounted, and the citation prompt asks only for insight ids.}
\label{fig:prompt-memory-controls}
\end{figure*}

\begin{figure*}[p]
\centering
\begin{solverbox}[ALFWorld solver system prompt]
Interact with a household to solve a task. Here are two examples.
<two worked episodes of the task's type from the ReAct paper, 40 lines>

Runtime notes:
- Play through the `act` tool. One call sends one command in the game's grammar and
  returns what the game says back, the same as the `> command` and observation pairs
  in the examples. Send one command per turn and read its observation before you send
  the next.
- The sandbox is offline and has no pip. You cannot install or download anything.
- The game ends the episode when the task is done or the budget runs out. You can also
  end it yourself by sending `done` through `act`, when you have finished or cannot
  finish. After that, `act` only answers `Episode over`.
\end{solverbox}
\begin{solverbox}[ALFWorld solver user prompt]
Here is the task.
You are in the middle of a room. Looking quickly around you, you see a bed 1, a diningtable 1, a drawer 4, a drawer 3, a drawer 2, a drawer 1, a garbagecan 1, a sidetable 2, and a sidetable 1.
Your task is to: put some book on sidetable.
\end{solverbox}
\caption{\textbf{No-memory solver prompt for ALFWorld, with the user prompt of one test task.} The instruction and exemplars follow the ReAct paper's ALFWorld loop.}
\label{fig:prompt-solver-alfworld}
\end{figure*}

\begin{figure*}[p]
\centering
\begin{solverbox}[ScienceWorld solver system prompt]
Interact with a household to solve a task. Here is an example.
<one worked episode of the task's type from SwiftSage's ReAct baseline, 76 lines>

Runtime notes:
- Play through the `act` tool. One call sends one command in the game's grammar and
  returns what the game says back, the same as the `> command` and observation pairs
  in the examples. Send one command per turn and read its observation before you send
  the next.
- The sandbox is offline and has no pip. You cannot install or download anything.
- The game ends the episode when the task is done or the budget runs out. You can also
  end it yourself by sending `done` through `act`, when you have finished or cannot
  finish. After that, `act` only answers `Episode over`.
\end{solverbox}
\begin{solverbox}[ScienceWorld solver user prompt]
Here is the task.
This outside location is called the outside. Here you see:   the agent  a substance called air  an axe  a fire pit (containing nothing)  a fountain (containing nothing)  the ground  a substance called wood You also see:  A door to the foundry (that is open)  A door to the greenhouse (that is open)  A door to the kitchen (that is open)
Your task is to boil caesium. For compounds without a boiling point, combusting the substance is also acceptable. First, focus on the substance. Then, take actions that will cause it to change its state of matter.
\end{solverbox}
\caption{\textbf{No-memory solver prompt for ScienceWorld, with the user prompt of one test task.} The instruction and exemplar follow SwiftSage's ReAct baseline, whose opening sentence is kept as it is.}
\label{fig:prompt-solver-scienceworld}
\end{figure*}

\begin{figure*}[p]
\centering
\begin{solverbox}[PlanCraft solver system prompt]
You are crafting in Minecraft. You need to decide on the next action.

Crafting Grid: The crafting table is organized into a 3x3 grid. Each slot in the grid has a unique identifier:
    - Top row: [A1] [A2] [A3]
    - Middle row: [B1] [B2] [B3]
    - Bottom row: [C1] [C2] [C3]

The output of the crafting process is placed in a designated output slot labeled [0] You cannot move or smelt items directly into slot [0]

Inventory Slots: The remaining inventory slots (outside of the crafting grid) are used for storing items. These slots are labeled as [I1] to [I36]

Actions:
	- move: Transfer a specific quantity of an item from one slot to another
	- smelt: Smelt an item in a furnace and moves the output to a specific slot
	- search: Search for recipes to craft a specific item
	- impossible: Stop task if it is certain that it is impossible with given inventory

Format
	- `move: from [Source] to [Target] with quantity N`
	- `smelt: from [Source] to [Target] with quantity N`
	- `search: <recipe name>`
	- `impossible: <reason>`

Example:
    - `move: from [I2] to [A1] with quantity 3`
    - `smelt: from [I5] to [I6] with quantity 1`

Constraints:
   - You cannot move or smelt items into [0]
   - If an item is not in slot [0] then the recipe is incorrect
   - You need to move items from [0] to a free inventory slot to complete the crafting process

<one worked episode from the authors' few-shot example, 42 lines>

Runtime notes:
- Play through the `take_action` tool. One call takes an `action` (move, smelt, search or
  impossible) and its `argument`, the text after the colon in the formats above, and
  returns what the game says back, the same as the `> command` and observation pairs in
  the example. Send one action per turn and read its observation before you send the next.
- The sandbox is offline and has no pip. You cannot install or download anything.
- The game ends the episode when the task is done or the budget runs out. You can also
  end it yourself with the `impossible` action, when you are certain the task cannot be
  done. After that, `take_action` only answers `Episode over`.
\end{solverbox}
\begin{solverbox}[PlanCraft solver user prompt]
Here is the task.
Craft an item of type: green_stained_glass
inventory:
 - sand [I1] quantity 8
 - birch_door [I5] quantity 54
 - tropical_fish [I12] quantity 39
 - green_glazed_terracotta [I17] quantity 62
 - purple_terracotta [I20] quantity 56
 - nether_quartz_ore [I22] quantity 62
 - dead_tube_coral_fan [I23] quantity 18
 - cactus [I27] quantity 1
 - blue_glazed_terracotta [I34] quantity 61
 - jungle_wood [I36] quantity 28
\end{solverbox}
\caption{\textbf{No-memory solver prompt for PlanCraft, with the user prompt of one test task.} The instruction and few-shot example are the authors' own.}
\label{fig:prompt-solver-plancraft}
\end{figure*}

\begin{figure*}[p]
\centering
\begin{solverdiffbox}[$\tau^3$-bench retail solver system prompt]
<instructions>
[-- You are a customer service agent that helps the user according to the <policy> provided below. In each turn you can either:-]
[++ You are a customer service agent that helps the user according to the <policy> provided below. Insights from past runs come with the task message, and the wiki at /work/corpus holds the runs behind them. In each turn you can either:+]
- Send a message to the user.
[-- - Make a tool call.-]
[-- You cannot do both at the same time. Try to be helpful and always follow the policy. Always make sure you generate valid JSON only.-]
[++ - Make a tool call to act on the store.+]
[++ - Read the wiki with the `read` or `bash` tool. The user does not see this.+]
[++ You cannot send a message and make a tool call at the same time. The insights can guide how you handle a request. Before relying on one, it helps to see how its source run used it. Try to be helpful and always follow the policy. Always make sure you generate valid JSON only.+]
</instructions>

<policy>
<the authors' retail policy, 136 lines, two of them changed:>
[-- At the beginning of the conversation, you have to authenticate the user identity by locating their user id via email, or via name + zip code. This has to be done even when the user already provides the user id.-]
[++ Do not reply to the user right away. First check the insights you will rely on in their source runs in the wiki, which the user does not see. A source run handled another request, even when it is the same user, so take from it how the policy was applied, never which order, item, address or payment method it chose. Ask this user. Then, before anything else in the conversation, you have to authenticate the user identity by locating their user id via email, or via name + zip code. This has to be done even when the user already provides the user id.+]
[-- You should not make up any information or knowledge or procedures not provided by the user or the tools, or give subjective recommendations or comments.-]
[++ You should not make up any information or knowledge, or give subjective recommendations or comments. Information comes only from the user and the tools. For how to handle a request, you may use the reusable procedural insights in the task message, after you check their evidence in the wiki and find that it fits this request and this policy.+]
</policy>

<available_tools>
Every action goes through the `take_action` tool, which takes `action` and `argument`. `action` is the name of one tool below, and `argument` is that tool's parameters as one JSON object.

<tool name="calculate">
Calculate the result of a mathematical expression.
parameters:
- expression (string, required): The mathematical expression to calculate, such as '2 + 2'. The expression can contain numbers, operators (+, -, *, /), parentheses, and spaces.
call: {"action": "calculate", "argument": {"expression": "<string>"}}
</tool>

<15 more retail tools in the same form>
</available_tools>

Runtime notes:
- Act on the store through the `take_action` tool. Its `action` is the name of one tool in
  <available_tools>, and its `argument` is what that tool's `call:` line shows. A call
  returns the tool's output or an `Error:` line. Send one action per turn and read its
  result before you send the next.
- The user's first message is in the task below. To talk to the user, reply with a message
  and no tool call. The user's answer comes back as the next message.
- The sandbox is offline and has no pip. You cannot install or download anything.
[-- The episode ends when the user ends the conversation, after ten rejected or failed tool calls, or when the step budget runs out. After that, `take_action` only answers `Episode over`.-]
[++ The episode ends when the user ends the conversation, after ten rejected or failed `take_action` calls, or when the step budget runs out. After that, `take_action` only answers `Episode over`.+]
\end{solverdiffbox}
\begin{solverbox}[$\tau^3$-bench retail solver user prompt]
Here is the task.
Hi, I’d like to cancel all of my pending orders, please.
\end{solverbox}
\caption{\textbf{No-memory solver prompt for $\tau^3$-bench retail, with the user prompt of one task.} The instruction and policy are the authors' own, and the 16 retail tools are rendered from their schemas. Red lines are removed and green lines added in WikiTrace's prompt.}
\label{fig:prompt-solver-tau3}
\end{figure*}

\begin{figure*}[p]
\centering
\begin{solverdiffbox}[AppWorld solver system prompt, part 1]
I am your supervisor, and you are an AI Assistant whose job is to complete my day-to-day tasks fully autonomously.

[-- To do this, you will need to interact with app(s) (e.g., spotify, venmo etc) using their associated APIs on my behalf. For this you will undertake a *multi-step conversation* using a python REPL environment. That is, you will write the python code, the environment will execute it and show you the result, based on which, you will write python code for the next step and so on, until you've achieved the goal. This environment will let you interact with app(s) using their associated APIs on my behalf.-]
[++ To do this, you will need to interact with app(s) (e.g., spotify, venmo etc) using their associated APIs on my behalf. For this you will undertake a *multi-step conversation* using a python REPL environment. That is, you will write the python code, the environment will execute it and show you the result, based on which, you will write python code for the next step and so on, until you've achieved the goal. This environment will let you interact with app(s) using their associated APIs on my behalf. Insights from past runs come with the task message, and the wiki at /work/corpus holds the runs behind them. The insights can guide how you solve a task. Before relying on one, it helps to see how its source run used it.+]
[+++]
[++ Before your first `exec`, check the insights you will rely on in their source runs in the wiki, with `read` or `bash`. A source run solved another task from another state, so take from it which APIs and steps worked, never its values or IDs. Then use the APIs below.+]

Here are three key APIs that you need to know to get more information

# To get a list of apps that are available to you.

```python
print(apis.api_docs.show_app_descriptions())
```

# To get the list of APIs under any app listed above, e.g. spotify

```python
print(apis.api_docs.show_api_descriptions(app_name='spotify'))
```

# To get the specification of a particular api, e.g. spotify app's login api

```python
print(apis.api_docs.show_api_doc(app_name='spotify', api_name='login'))
```

Each code execution will produce an output that you can use in subsequent calls. Using these APIs, you can now generate code, that I will execute, to solve the task. Let's start with the task

<a worked example of another task with its task header, 313 lines, in which one line is added before the first exec:>
[++ [read / bash: checking the source runs of the insights to rely on, omitted]+]
\end{solverdiffbox}
\caption{\textbf{No-memory solver system prompt for AppWorld, first part.} The instruction and worked example are those of AppWorld's ReAct code agent. Red lines are removed and green lines added in WikiTrace's prompt.}
\label{fig:prompt-solver-appworld}
\end{figure*}

\begin{figure*}[p]
\centering
\begin{solverdiffbox}[AppWorld solver system prompt, part 2]
**Key instructions**:

A. General instructions:

- Act fully on your own. You must make all decisions yourself and never ask me or anyone else to confirm or clarify. Your role is to solve the task, not to bounce questions back, or provide me directions to follow.
- You have full access -- complete permission to operate across my connected accounts and services.
- Never invent or guess values. For example, if I ask you to play a song, do not assume the ID is 123. Instead, look it up properly through the right API.
[++ - For how to solve a task, you may use the insights in the task message, after you check their evidence in the wiki and find that it fits this task.+]
- Never leave placeholders; don't output things like "your_username". Always fill in the real value by retrieving it via APIs (e.g., Supervisor app for credentials).
- When I omit details, choose any valid value. For example, if I ask you to buy something but don't specify which payment card to use, you may pick any one of my available cards.
- Avoid collateral damage. Only perform what I explicitly ask for. Example: if I ask you to buy something, do not delete emails, return the order, or perform unrelated account operations.

B. App-specific instructions:

- All my personal information (biographical details, credentials, addresses, cards) is stored in the Supervisor app, accessible via its APIs.
- Any reference to my friends, family or any other person or relation refers to the people in my phone's contacts list.
- Always obtain the current date or time, from Python function calls like `datetime.now()`, or from the phone app's get_current_date_and_time API, never from your internal clock.
- All requests are concerning a single, default (no) time zone.
- For temporal requests, use proper time boundaries, e.g., when asked about periods like "yesterday", use complete ranges: 00:00:00 to 23:59:59.
- References to "file system" mean the file system app, not the machine's OS. Do not use OS modules or functions.
- Paginated APIs: Always process all results, looping through the page_index. Don't stop at the first page.

C. Code-operation instructions

- Send each chunk of code as one `exec` call. Code written in your reply text does not run.
- Remember, you can use the variables in your code in subsequent `exec` calls.
[-- - Remember that the email addresses, access tokens and variables (e.g. spotify_password) in the example above are not valid anymore.-]
[-- - Always look at API specifications (using apis.api_docs.show_api_doc) before calling an API.-]
[++ - Remember that the email addresses, access tokens and variables (e.g. spotify_password) in the example above and in the past runs in the wiki are not valid anymore.+]
[++ - Do not look up API specifications right away. First check the insights you will rely on in their source runs in the wiki, with `read` or `bash`. Then always look at API specifications (using apis.api_docs.show_api_doc) before calling an API.+]
- Write small chunks of code and only one chunk of code in every `exec` call. Make sure everything is working correctly before making any irreversible changes.
- The Python environment supports the standard library. But system-level operations that may access or affect OS files, processes, etc., are not allowed and will raise an error if called.
- To interact with apps, only use the provided app APIs, and not the corresponding Python packages, e.g., do NOT use `spotipy` for Spotify.
- The provided API documentation has both the input arguments and the output JSON format. Use this information when making API calls and parsing their outputs.

D. Task-completion instructions:

You must call the `apis.supervisor.complete_task` API after completing the task.
- If an answer is needed, e.g., for "How many songs are in the Spotify queue?", call it with the appropriate answer argument value.
- If no answer is required, e.g., for "Start my Spotify music player.", omit the answer argument (or set it to None/null).
- The task is doable, but if you cannot find a way, you can call it with status="fail" to exit with failure.

When the answer is given:
- Keep answers minimal. Return only the entity, number, or direct value requested - not full sentences.
  E.g., for the song title of the current playing track, return just the title.
- Numbers must be numeric and not in words.
  E.g., for the number of songs in the queue, return "10", not "ten".

Runtime notes:
- Each `> exec` in the example is one call of the `exec` tool. Send the code as that
  call, never as reply text, and read its `output:` before the next call. Variables
  persist between calls. `exec` is the only way to reach the apps. Running python3
  from bash does not work.
- The sandbox is offline and has no pip. You cannot install or download anything.
- The episode ends when you call `apis.supervisor.complete_task(...)`, or when the exec
  budget runs out. Call it with `status="fail"` if you cannot finish. After that, `exec`
  only answers `Episode over`.
\end{solverdiffbox}
\begin{solverbox}[AppWorld solver user prompt]
Using these APIs, now generate code to solve the actual task:

My name is: Katherine Smith. My personal email is ka_ball@gmail.com and phone number is 7863159797.
Task: I jotted down some songs in Simple Note recently. Make a playlist titled "Random Songs" out of it.
\end{solverbox}
\caption{\textbf{No-memory solver prompt for AppWorld, second part, with the user prompt of one test task.} The key instructions are those of AppWorld's ReAct code agent. Red lines are removed and green lines added in WikiTrace's prompt.}
\label{fig:prompt-solver-appworld-2}
\end{figure*}

\end{document}